\documentclass{article}

\usepackage{arxiv}

\usepackage[utf8]{inputenc} % allow utf-8 input
\usepackage[T1]{fontenc}    % use 8-bit T1 fonts
\usepackage{hyperref}       % hyperlinks
\usepackage{url}            % simple URL typesetting
\usepackage{booktabs}       % professional-quality tables
\usepackage{amsfonts}       % blackboard math symbols
\usepackage{nicefrac}       % compact symbols for 1/2, etc.
\usepackage{lipsum}
\usepackage{graphicx}
\usepackage{cite}
\usepackage{cite,amsmath,amssymb,amsfonts,algorithmic,graphicx}
\usepackage{url,enumitem,bm}
\usepackage{algorithmic}
\usepackage{graphicx}
\usepackage{textcomp}
\usepackage{subfigure}
\usepackage{xcolor}
\usepackage{soul, color, xcolor}
\usepackage{amsmath}
\usepackage{graphicx}
\usepackage{multirow}
\usepackage{booktabs}
\usepackage{amssymb}
\usepackage{pifont}   
\usepackage{hyperref}
\usepackage[table]{xcolor}
\usepackage{todonotes}
\usepackage[commandnameprefix=ifneeded]{changes}
\definecolor{deepgreen}{rgb}{0.0, 0.5, 0.0} % 定义一个深绿色
\definechangesauthor[name={Reviewer 1}, color=blue]{R1}
\definechangesauthor[name={Reviewer 2}, color=red]{R2}
\definechangesauthor[name={Reviewer 3}, color=deepgreen]{R3}

\newcommand{\xmark}{\ding{55}}
\usepackage[ruled]{algorithm2e}
\def\BibTeX{{\rm B\kern-.05em{\sc i\kern-.025em b}\kern-.08em
    T\kern-.1667em\lower.7ex\hbox{E}\kern-.125emX}}
\begin{document}

\title{Heterogeneous Vertiport Selection Optimization for On-Demand Air Taxi Services: A Deep Reinforcement Learning Approach}

\author{
Aoyu~Pang,
Maonan~Wang,
Zifan~Sha,
Wenwei~Yue,
Changle~Li,
Chung~Shue~Chen,
and Man-On~Pun
\\[0.5em]
\small
Aoyu~Pang and Man-On~Pun are with the School of Science and Engineering,\\
The Chinese University of Hong Kong, Shenzhen, China.\\
Maonan~Wang is with the School of Science and Engineering,\\
The Chinese University of Hong Kong, Shenzhen, China, and the Shanghai AI Laboratory, Shanghai, China.\\
Zifan~Sha, Wenwei~Yue, and Changle~Li are with the State Key Laboratory of Integrated Services Networks,\\
Xidian University, Xi'an, China.\\
Chung~Shue~Chen is with Nokia Bell Labs, Paris-Saclay, France.
}

\maketitle
\begin{abstract}
Urban Air Mobility (UAM) has emerged as a transformative solution to alleviate urban congestion by utilizing low-altitude airspace, thereby reducing pressure on ground transportation networks. To enable truly efficient and seamless door-to-door travel experiences, UAM requires close integration with existing ground transportation infrastructure. However, current research on optimal integrated routing strategies for passengers in air-ground mobility systems remains limited, with a lack of systematic exploration.To address this gap, we first propose a unified optimization model that integrates strategy selection for both air and ground transportation. This model captures the dynamic characteristics of multimodal transport networks and incorporates real-time traffic conditions alongside passenger decision-making behavior. Building on this model, we propose a Unified Air-Ground Mobility Coordination (UAGMC) framework, which leverages deep reinforcement learning (RL) and Vehicle-to-Everything (V2X) communication to optimize vertiport selection and dynamically plan air taxi routes. Experimental results demonstrate that UAGMC achieves a 34\% reduction in average travel time compared to conventional proportional allocation methods, enhancing overall travel efficiency and providing novel insights into the integration and optimization of multimodal transportation systems. This work lays a solid foundation for advancing intelligent urban mobility solutions through the coordination of air and ground transportation modes. The related code can be found at \url{https://github.com/Traffic-Alpha/UAGMC}.
\end{abstract}

\section{Introduction} \label{sec:introduction}

The rapid acceleration of global urbanization has concentrated economic and social activities within metropolitan areas, giving rise to a series of urgent challenges, including severe traffic congestion, environmental degradation, and overpopulation~\cite{RN16}. For instance, it has been reported that $30\%$ of the population spends over 60 minutes commuting daily~\cite{TrafficReport, yan2024urban}. This situation highlights the pressing need for innovative solutions to enhance urban transportation capacity. While existing traffic optimization methods have attempted to improve the efficiency of ground transportation~\cite{wang2024llm, pang2024scalable}, they have yet to explore the potential of low-altitude airspace as a complementary resource.

In this context, the emergence of Urban Air Mobility (UAM) represents a promising   shift. By leveraging underutilized low-altitude airspace, UAM offers a sustainable solution to longstanding urban traffic congestion issues~\cite{pons2022understanding}. Leading organizations, including Uber~\cite{elevate2016fast}, Ehang~\cite{mittal2023design}, and NASA~\cite{johnson2022nasa}, have initiated studies and pilot programs to explore UAM’s potential. UAM fundamentally transforms urban transportation by expanding traditional two-dimensional (2D) systems into three-dimensional (3D) networks~\cite{song2021development}, offering innovative and scalable strategies for mitigating urban congestion.

\begin{figure}[t]
    \centering
    \includegraphics[width=0.5\linewidth]{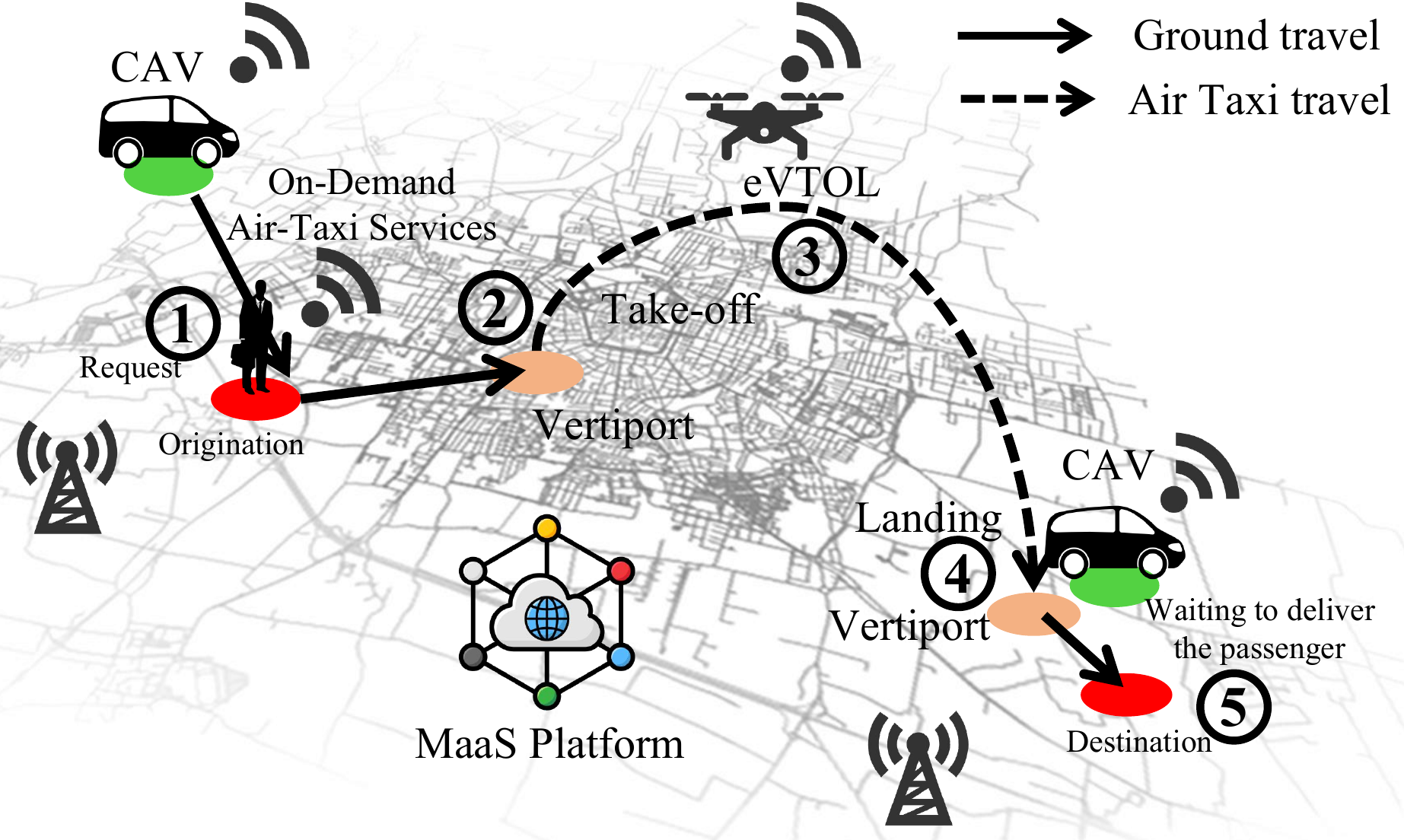}
    \caption{Overall system model of door-to-door air taxi service.}
    \label{fig:framework}
\end{figure}

Among its many benefits, UAM has been shown to significantly reduce travel time through the deployment of highly maneuverable, on-demand air mobility services~\cite{rothfeld2021potential}. Air taxis are a crucial element of UAM, powered by electric vertical take-off and landing (eVTOL) vehicles. These vehicles provide innovative solutions to urban challenges by reducing traffic congestion, lowering carbon emissions, and improving urban connectivity. Around the world, cities are actively advancing UAM initiatives: Dubai is preparing to launch air taxi services for tourists, Singapore is developing a commercial route connecting Marina Bay and Sentosa, and Shenzhen, China, is collaborating with enterprises to establish its UAM system. With their ability to deliver faster, more efficient, and environmentally friendly transportation, air taxis are poised to transform urban mobility and outperform traditional ground transportation systems~\cite{haan2021commuter, ilahi2021understanding}.

Air taxi services represent a quintessential multimodal transportation solution, seamlessly integrating aerial and ground travel to deliver a comprehensive door-to-door mobility experience~\cite{boddupalli2024mode}. As depicted in Fig.~\ref{fig:framework}, passengers initiate an air taxi service request via a Mobility-as-a-Service (MaaS) platform, which employs Vehicle-to-Everything (V2X) network data to evaluate real-time conditions, including ground traffic status, vertiport capacity, and the availability of eVTOL vehicles ~\cite{stopka2018mobility, rajendran2020air}. Based on this analysis, the system assigns a connected autonomous vehicle (CAV) for ground transportation while selecting the optimal departure and arrival vertiports. Passengers are first transported by the assigned CAV to the designated vertiport, where they briefly wait to board the air taxi. The air taxi then navigates predetermined low-altitude flight corridors to bypass ground traffic congestion, efficiently traveling to the destination vertiport. Upon arrival, passengers transition seamlessly to pre-arranged ground transportation for the final leg of their journey. This fully integrated system effectively addresses the first- and last-mile connectivity challenges, enhancing the overall efficiency and convenience of UAM solutions \cite{rajendran2020air}.

eVTOLs are inherently limited in passenger capacity and rely on fixed vertiports, which significantly constrain the scalability of UAM services. Imbalances in the spatial distribution and service capacity of vertiports often lead to congestion and long waiting times at popular locations, while other vertiports remain underutilized.
In highly dynamic urban environments, continuous fluctuations in passenger demand distribution and vehicle availability further exacerbate the complexity of system scheduling and decision-making, thereby adversely affecting overall travel efficiency. Previous studies have explored various dimensions of UAM, including the strategic planning of vertiport locations with particular attention to land use constraints~\cite{yedavalli2022planning, zhou2020optimized, zhao2024strategic}, the optimization of air taxi route design~\cite{wu2021integrated}, and the modeling and forecasting of passenger demand~\cite{bulusu2021traffic}. However, the optimization of end-to-end air taxi travel strategies, especially those involving the coordination of ground and aerial transportation, remains insufficiently explored. Unlike traditional multimodal systems such as subways or buses~\cite{ilahi2021understanding, auad2022ridesharing, chen2022transfer}, individual eVTOLs typically carry only 2 to 5 passengers. In the absence of a unified decision-making framework, passengers tend to choose the nearest vertiport, often overlooking slightly farther but less congested alternatives. This behavior further exacerbates resource imbalance and increases overall travel times. 

Traditional optimization methods rely on static modeling and thus struggle to adapt to real-time changes in system states, exhibiting clear limitations in adaptability ~\cite{nazari2018reinforcement}. In contrast, reinforcement learning (RL) excels at handling high-dimensional and uncertain environments by enabling continuous policy updates through ongoing interactions. In particular, the Actor-Critic algorithm, which integrates policy optimization with value estimation, achieves stable and efficient learning in large-scale state-action spaces, offering a promising solution for real-time coordination in integrated air-ground mobility systems ~\cite{jali2024efficient, ling2023deep}. 

Building on these insights, we propose a framework called \textbf{U}nified \textbf{A}ir-\textbf{G}round \textbf{M}obility \textbf{C}oordination Optimization (UAGMC). This framework aims to optimize end-to-end air taxi travel strategies by integrating both ground and air transportation within a unified decision-making mechanism. By integrating deep learning techniques, the UAGMC framework can extract valuable information from extensive traffic data and optimize passenger routing strategies in dynamic traffic environments, significantly improving overall transportation efficiency and reducing travel time. The framework seamlessly integrates ground and air transportation systems, coordinating passenger movement from origin to destination, facilitating efficient transfers, and minimizing overall travel time while enhancing transportation efficiency. The main contributions can be summarized as follows:

% 重新强调一下 贡献点
\begin{itemize}[leftmargin=*]
\item  To the best of our knowledge, this is the {\em first} attempt to explore strategy planning that integrates air taxis with ground transportation. We simultaneously consider ground and air transportation to establish a door-to-door air taxi route planning model, formulating it as a Markov Decision Process (MDP) problem. This framework integrates CAV with air taxi services, optimizing passenger vertiport selection through the integration of multimodal transportation, significantly enhancing overall travel efficiency.

\item To address the heterogeneous optimization problem of vertiport selection, we propose the UAGMC framework. The framework leverages deep RL with states and actions designed to address both the heterogeneity issue and the sparse reward problem in door-to-door air taxi services. Moreover, we design Multi-Source Contextual Embedding (MSCE)  and the Spatio-Temporal Integration Network (STIN)  that effectively integrate and extract key features from multi-source heterogeneous information. This enhances the model’s ability to represent high-dimensional complex data, significantly improving the learning efficiency and overall performance of the framework.

\item Extensive experimental evaluations demonstrate that UAGMC consistently outperforms rule-based, queuing-theoretic, and conventional RL baselines, achieving significant reductions in overall travel time. 
Further analysis shows that these gains mainly arise from congestion-aware vertiport selection, and confirms the necessity of MSCE and STIN as well as the robustness of UAGMC under varying demand and vehicle capacity settings.

\end{itemize}

The rest of the paper is organized as follows. Section~\ref{sec:related_works} provides a review of the relevant literature while Section~\ref{sec:problem formulation} defines the traffic terminology pertinent to our discussion. After that, the core of our proposed methodology, including the historical-future data fusion and action refinement modules, is detailed in Section~\ref{sec:framework}. Section~\ref{sec:experiment} describes the experimental framework, outlines the benchmark methods, and evaluates the performance of our proposed framework against these benchmarks with a focus on the average waiting time of vehicles. Finally, Section~\ref{sec:conclusion} concludes our findings and offers insights into potential avenues for future research in this domain.

\section{Related Works} \label{sec:related_works}
\subsection{Urban Air Mobility}
As a vital complement to ground transportation, UAM is rapidly emerging as a key technology. Current research addresses a range of key issues, including public acceptance, demand forecasting, infrastructure design, flight route planning, and its competitive potential compared to traditional transportation modes. For example, the establishment of Air Traffic Management (ATM) systems~\cite{wang2023quasi} ensures the safe operation of multiple eVTOL vehicles in urban airspaces. Additionally, UAM application scenarios have expanded to include air taxis, airport shuttles, and intercity flights~\cite{balac2019prospects, rajendran2020air}, with air taxis attracting significant attention due to their on-demand service capabilities~\cite{haan2021commuter}.

Air taxis utilize small eVTOL vehicles to connect multiple vertiports, enabling rapid and efficient urban transportation while alleviating ground traffic congestion. By leveraging the eVTOL concept, air taxis gain operational advantages through the use of dedicated vertiports. However, despite their potential, air taxis face significant technical challenges. Current research focuses on noise control, strategies for aircraft charging and energy replenishment~\cite{garrow2021urban, zou2023imbalance}, and vertiport layout optimization~\cite{zhou2020optimized, wu2021integrated}. These efforts provide foundational support for the initial deployment of UAM systems, yet the deep integration of air and ground transportation systems remains underexplored.

Achieving truly seamless door-to-door air taxi services relies on the effective integration of air taxis, autonomous vehicles, and public transit systems to enable coordinated operations in a multimodal transportation network~\cite{yan2024urban, babic2022integrated}. While considerable progress has been made in optimizing ground transportation, such as autonomous vehicle matching~\cite{qin2021optimizing} and dynamic route planning~\cite{ben2012dynamic, batista2021dynamic}, research on coordination between air and ground transportation remains limited. Existing studies have shown that such coordination significantly impacts overall travel efficiency~\cite{bennaceur2022passenger, you2024toward}, and that passenger demand for UAM services is highly sensitive to vertiport access time and boarding alighting delays~\cite{boddupalli2024mode}, further emphasizing the critical role of ground accessibility in system performance. Despite growing recognition of the importance of air-ground integration, most current research remains focused on vertiport siting and network-level optimization~\cite{jin2024robust, rajendran2020air, zhao2024strategic}, with limited attention to modeling and analyzing how passengers should select among multiple candidate vertiports under realistic ground access conditions.

To bridge these gaps, this paper proposes an optimization framework for passenger travel strategies in the air taxi multimodal network. By integrating vertiport selection and route planning, the framework aims to maximize the overall operational efficiency of UAM systems and facilitate seamless integration between air and ground transportation.

\subsection{Heterogeneous vertiports system option}

Vertiport selection in interconnected air taxi systems is a challenging heterogeneous optimization problem influenced by varying service rates, locations, and operational constraints. Optimizing such heterogeneous service systems, characterized by diverse capacities and dynamics, remains a long-standing problem in operations research. Early studies~\cite{lin1984optimal, walrand1984note} established the optimality of threshold policies in simplified two-server settings, with extensions via value iteration methods~\cite{koole1995simple}. However, generalizing these results to large-scale multi-server heterogeneous systems remains unresolved due to intrinsic interdependencies among service rates, arrivals, and system states~\cite{koole2022slow, rykov2001monotone, de2006incomplete}. 

In response to the limitations of analytical methods, data-driven approaches, particularly RL, have emerged as promising tools for optimizing complex service systems~\cite{jali2024efficient}. RL has demonstrated its effectiveness across various domains, including queueing networks~\cite{wei2024sample}, inventory control~\cite{mannor2003cross}, and traffic management~\cite{pang2024scalable}. Recent advances focus on scaling RL to high-dimensional, heterogeneous systems, addressing challenges such as infinite state spaces while ensuring stability and convergence. For instance, \cite{dai2022queueing} utilized policy gradient algorithms like PPO and TRPO, incorporating Lyapunov stability arguments to handle unbounded state and cost spaces. Similarly, \cite{liu2019reinforcement} introduced truncation-based techniques to maintain optimality in large-scale systems.

RL provides a robust framework for tackling the complexities of heterogeneous vertiport systems. Parameterized by both policies (actors) and value functions (critics), the Actor-Critic algorithm effectively balances sample efficiency with convergence guarantees, making it suitable for large-scale heterogeneous environments. By leveraging real-time operational data, RL dynamically allocates air taxis to vertiports based on current demand, vertiport availability, and projected waiting times, outperforming static or heuristic approaches. RL-driven strategies have already shown promise in task allocation~\cite{staffolani2023rlq} and distributed scheduling, adapting to dynamic environments while ensuring efficient resource utilization.

This paper presents a novel RL-based optimization framework specifically designed for heterogeneous vertiport systems. The proposed framework integrates unique structural constraints and operational dynamics in UAM, utilizing policy gradient methods to achieve scalability and stability. By addressing the complex interdependencies among vertiport heterogeneity, operational demands, and real-time decision-making, this framework represents a significant advancement in optimizing UAM networks, bridging theoretical innovation and practical application.

\subsection{Reinforcement Learning for Complex Environment} 
In recent years, RL has garnered significant attention, focusing primarily on enhancing agents' decision-making capabilities in dynamic and complex environments. Common RL algorithms include value-based methods, policy gradient methods, and Actor-Critic frameworks~\cite{sutton2018reinforcement}. Among them, Q-Learning, a classical value-based approach, derives the optimal policy by learning the state-action value function. Q-Learning improves over traditional methods with richer state representations but remains limited in scalability and optimality in high-dimensional, complex environments~\cite{li2016traffic}.

To overcome these limitations, Deep RL integrates the strengths of Deep Learning (DL) and RL by leveraging deep neural networks to model high-dimensional state spaces, significantly improving performance. For example, Deep Q-Networks (DQNs) utilize deep models to learn Q-functions, finding extensive applications in fields such as path planning~\cite{peng2021urban}, traffic scheduling~\cite{qin2021optimizing}, and multi-agent system coordination. \cite{rajendran2020air} applied deep reinforcement learning (DRL) to air taxi dispatching in complex urban environments, significantly enhancing traffic efficiency. Moreover, some studies have further optimized state space representations. For instance, \cite{wang2024unitsa} proposed a matrix-based representation for traffic states, while \cite{pang2024scalable} and \cite{shabestray2019multimodal} employed Convolutional Neural Networks (CNNs) to process image-like vehicle position data, thereby enhancing the understanding of traffic environments. With the rapid advancement of autonomous driving and connected vehicle technologies, upgraded sensors and infrastructure further improve traffic data collection and processing capabilities~\cite{li2018piecewise}. For example, \cite{chu2019multi} leveraged Long Short-Term Memory (LSTM) networks to process diverse state inputs, providing effective solutions for dynamic and complex environments.  In recent years, Transformer architectures based on self-attention mechanisms have been widely adopted in reinforcement learning and intelligent transportation systems due to their strong capability to efficiently model global temporal dependencies in high-dimensional, long-horizon decision-making problems, demonstrating improved stability and generalization performance~\cite{yu2025hybrid}.

In the domain of combinatorial optimization, such as the Traveling Salesman Problem (TSP), DRL has exhibited exceptional problem-solving capabilities~\cite{zhang2022deep}. Many methods are based on encoder-decoder frameworks~\cite{zhang2021solving}, which achieve objectives by optimizing long-term rewards, such as minimizing the total travel distance in dynamic TSP. Typical approaches include policy gradient methods, Q-Learning~\cite{mnih2015human} and its extensions, and Actor-Critic algorithms~\cite{nazari2018reinforcement}. For instance, \cite{nazari2018reinforcement} proposed an unsupervised network for online vehicle routing, ~\cite{james2019online} developed a time-aware online Q-Learning algorithm, and ~\cite{lu2019learning} incorporated heuristic search into the DRL framework to optimize initial path planning solutions, offering novel approaches for dynamic path planning in complex environments.

To address the limitations of traditional RL methods in handling high-dimensional heterogeneous features, this paper proposes an improved DRL algorithm. The approach optimizes decision-making efficiency through the synergy between the actor and critic components. Furthermore, the framework incorporates an encoder and LSTM networks to process diverse state variables, significantly enhancing system performance and operational efficiency.

\begin{figure}[ht]
    \centering
    \includegraphics[width=0.6\linewidth]{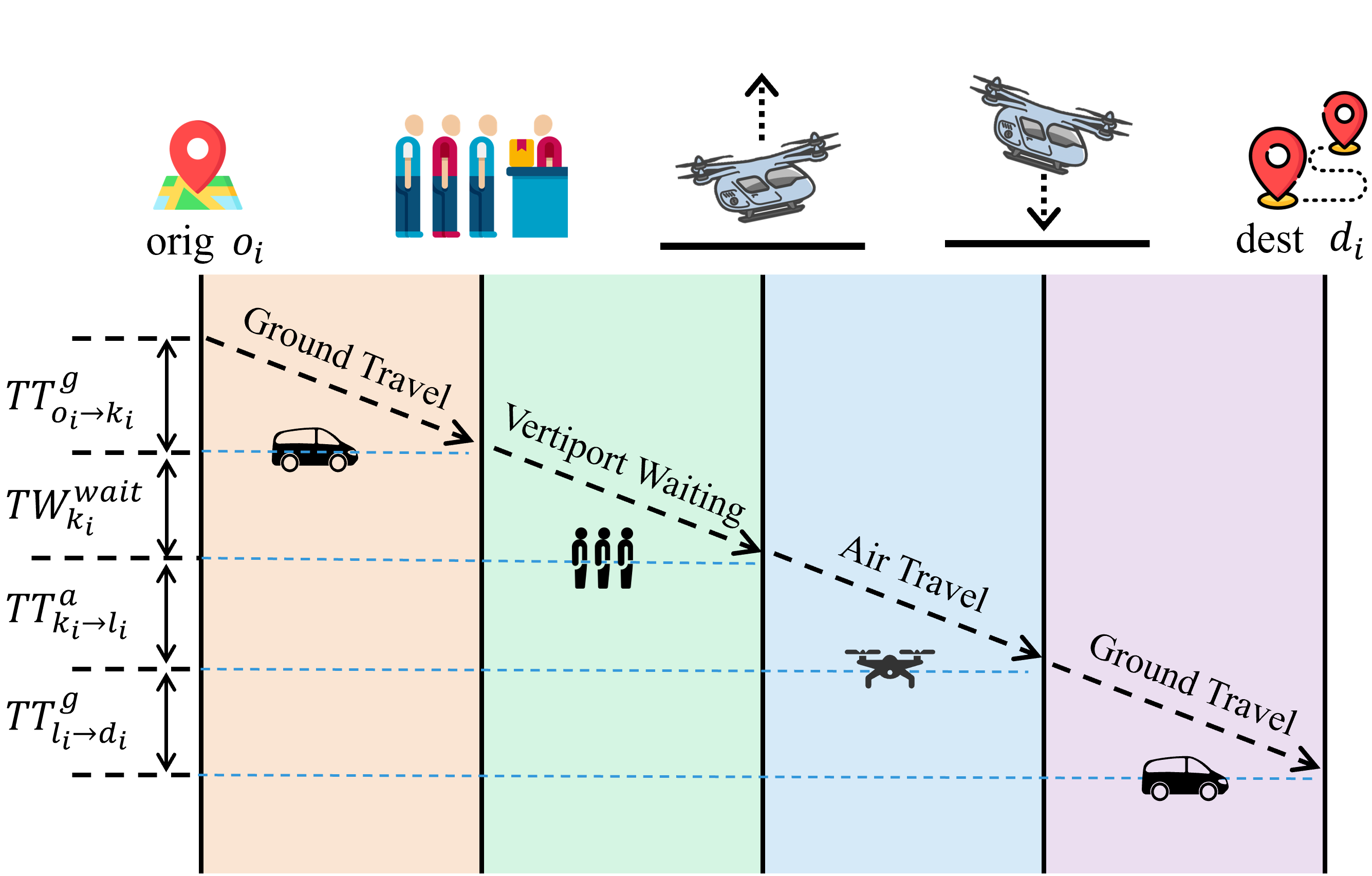}
    \caption{Illustration of the door-to-door air taxi service process with four-stage.}
    \label{fig:problem_defintion}
\end{figure}

\begin{table}[ht]
\centering
\caption{Variable Definitions for Multimodal Air-Ground Passenger Routing (All variables defined for passenger $i$ unless stated).}
\begin{tabular}{ll}
    \toprule
    \textbf{Variable} & \textbf{Definition} \\
    \midrule
    % Basic Sets
    \multicolumn{2}{l}{\textbf{\textit{1. Basic Sets and Passenger Definitions}}} \\
    \midrule
    $\mathcal{N}$ & Set of all passengers \\
    $T$ & Time horizon for system optimization \\
    $\mathcal{N}(t)$ & Passengers appeared by time $t$, $\sum_{t=1}^{T}{\mathcal{N}(t)}=\mathcal{N}$ \\
    $\mathcal{V}$ & Set of vertiports \\
    $o_i$, $d_i$ & Origin and destination of passenger $i$ \\
    % Node & Path
    \midrule
    \multicolumn{2}{l}{\textbf{\textit{2. Vertiport Paths and Distances}}} \\
    \midrule
    $k_{i}, l_{i} \in \mathcal{V}$ & Departure and arrival vertiports \\
    $L^{\text{g}}_{o_i \rightarrow k_i}$ & Ground distance: origin $o_i$ to $k_i$ \\
    $L^{\text{g}}_{l_i \rightarrow d_i}$ & Ground distance: $l_i$ to destination $d_i$ \\
    $L^{\text{a}}_{k_i \rightarrow l_i}$ & Air distance: from $k_i$ to $l_i$ \\
    % Ground Travel
    \midrule
    \multicolumn{2}{l}{\textbf{\textit{3. Ground Travel}}} \\
    \midrule
    $TT^{\text{g}}_{o_i \rightarrow k_i}$ & Travel time from $o_i$ to vertiport $k_i$ \\
    $TT^{\text{g}}_{l_i \rightarrow d_i}$ & Travel time from vertiport $l_i$ to $d_i$ \\
    $\bar{v}^{\text{g}}_{o_i \rightarrow k_i}, \bar{v}^{\text{g}}_{l_i \rightarrow d_i}$ & Average speed from $o_i$ to $k_i$ and $l_i$ to $d_i$ \\
    % Air Travel
    \midrule
    \multicolumn{2}{l}{\textbf{\textit{4. Air Travel (eVTOL between vertiports)}}} \\
    \midrule
    $\bar{v}^{\text{a}}_{k_i \rightarrow l_i}$ & Average airspeed from $k_i$ to $l_i$ \\
    $TT^{\text{a}}_{k_i \rightarrow l_i}$ & Air travel time from $k_i$ to $l_i$ \\
    $TT^{\epsilon}$ & Constant delay (takeoff/landing) \\
    % Waiting and Service Modeling
    \midrule
    \multicolumn{2}{l}{\textbf{\textit{5. Waiting and Service at Vertiport $k$}}} \\
    \midrule
    $q^{\text{wait}}_{k_i}(t)$ & Queue length at time $t$ \\
    $ \bar v^{\text{serv}}_{k_i}$ & Service rate of vertiport $k$ \\
    $N^{\text{arr}}_{k_i}(t)$ & Number of new arrivals at time $t$ \\
    $N^{\text{dep}}_{k_i}(t)$ & Number of eVTOL departures at time $t$ \\
    $C_{k_i}$ & Passenger capacity of each eVTOL \\
    $TW^{\text{wait}}_{k_i}$ & Waiting time at vertiport $k$ \\
    % Total Travel Time
    \midrule
    \multicolumn{2}{l}{\textbf{\textit{6. Total Travel Time}}} \\
    \midrule
    $TT^{\text{tot}}_{i, k, l}$ & Total travel time from $o_i$ to $d_i$ via $k_i \rightarrow l_i$ \\
    \bottomrule
\end{tabular}
\label{tab:variable_definitions}
\end{table}

% 对于变量命名的表格重写，字母定义更加清晰, 表格加入了分区
% 加入 subsection，逻辑清晰，行驶时间由若干个部分组成
% 重新绘制了图片, 时间包含四个部分
\section{Problem Formulation} \label{sec:problem formulation}

We consider a multimodal passenger transport system integrating ground-based CAVs with UAM services via eVTOLs. The objective is to minimize the total travel time $TT^{\text{tot}}_{i,k,l}$ for all passengers over a fixed time horizon $T$, by jointly optimizing their routing decisions across the ground-air-ground travel chain.

Each passenger $i \in \mathcal{N}$ is associated with an origin location $o_i$ and a destination $d_i$. As shown in Fig.~\ref{fig:problem_defintion}, a passenger's journey consists of four segments: (1) ground travel from origin $o_i$ to a departure vertiport $k_{i} \in \mathcal{V}$, (2) waiting at vertiport $k_{i}$, (3) air travel from vertiport $k_{i}$ to a landing vertiport $l_{i} \in \mathcal{V}$, and (4) ground travel from $l_{i}$ to $d_i$. The goal is to minimize the total travel time by jointly optimizing the departure and landing vertiport pair $(k_{i}, l_{i})$ for each passenger $i$. The total travel time $TT^{\text{tot}}_{o_i, k_i, l_i, d_i}$ is formulated as:
\begin{equation} \label{eq:total_tt}
    TT^{\text{tot}}_{o_i, k_i, l_i, d_i} 
    = 
    TT^{\text{g}}_{o_i \rightarrow k_i} 
        + TW^{\text{wait}}_{k_i} 
        + TT^{\text{a}}_{k_i \rightarrow l_i} 
        + TT^{\text{g}}_{l_i \rightarrow d_i} 
        + TT^{\epsilon},
\end{equation}
where $TT^{\text{g}}_{o_i \rightarrow k_i}$ denotes the ground-based travel time from origin $o_i$ to departure vertiport $k_i$. $TW^{\text{wait}}_{k_i}$ denotes the estimated waiting time at vertiport $k_i$, corresponding to the duration a passenger must wait for boarding and departure after arriving at the vertiport. $TT^{\text{a}}_{k_i \rightarrow l_i}$ is the aerial travel time between vertiports $k_i$ and $l_i$. $TT^{\text{g}}_{l_i \rightarrow d_i}$ denotes the final ground travel time from arrival vertiport $l_i$ to destination $d_i$, and $TT^{\epsilon}$ captures a fixed overhead time accounting for vertical takeoff and landing operations. 

Notably, pre-request response delays and CAV dispatch times prior to pickup are excluded from this formulation, as these are independent of vertiport assignment and thus do not influence the optimization outcome. The detailed modeling and computation of each component in Eq.~\eqref{eq:total_tt} will be presented in the following subsections. Furthermore, a complete list of all variables and their definitions can be found in Table~\ref{tab:variable_definitions}.

% Ground Travel Time
\subsection{Ground Travel Time}

The ground travel time consists of two segments: (i) from the passenger's origin $o_i$ to the selected departure vertiport $k_i$, and (ii) from the arrival vertiport $l_i$ to the passenger’s final destination $d_i$. Specifically, the ground travel time from $o_i$ to $k_i$ is given by:
\begin{equation}
    TT^{\text{g}}_{o_i \rightarrow k_i} 
    = 
    \frac{L^{\text{g}}_{o_i \rightarrow k_i}}{\bar{v}^{\text{g}}_{o_i \rightarrow k_i}},
\end{equation}
where $L^{\text{g}}_{o_i \rightarrow k_i}$ denotes the shortest feasible travel distance from origin $o_i$ to vertiport $k_i$, and $\bar{v}^{\text{g}}_{o_i \rightarrow k_i}$ is the estimated average travel speed along the selected path. The average speed is estimated using the Greenberg traffic flow model \cite{greenberg1959analysis}, which is commonly used for urban traffic conditions and is:
\begin{equation} \label{eq:greenberg_model}
    \bar{v} = v_m \ln\left(\frac{k_j}{k_c}\right),
\end{equation}
where $v_m$ denotes the speed corresponding to maximum traffic flow, $k_c$ is the current traffic density, and $k_j$ represents the jam density of the road segment. Similarly, the ground travel time from the arrival vertiport $l_i$ to the destination $d_i$ is computed as:
\begin{equation}
    TT^{\text{g}}_{l_i \rightarrow d_i} 
    = 
    \frac{L^{\text{g}}_{l_i \rightarrow d_i}}{\bar{v}^{\text{g}}_{l_i \rightarrow d_i}}.
\end{equation}
where all terms are defined analogously. The shortest-distance paths $L^{\text{g}}$ are obtained via $A^{*}$ algorithm \cite{candra2020dijkstra}, which is widely employed in intelligent transportation systems and urban traffic navigation applications.

% Air Travel Time
\subsection{Air Travel Time}

The air travel time between vertiports $k_i$ and $l_i$ is:
\begin{equation}
    TT^{\text{a}}_{k_i \rightarrow l_i} 
    = 
    \frac{L^{\text{a}}_{k_i \rightarrow l_i}}{\bar{v}^{\text{a}}_{k_i \rightarrow l_i}},
\end{equation}
where $L^{\text{a}}_{k_i \rightarrow l_i}$ is the air distance and $\bar{v}^{\text{a}}_{k_i \rightarrow l_i}$ is the average eVTOL flight speed. In contrast to ground speed, the flight speed is assumed to be stable and unaffected by congestion. All air routes are pre-validated to be within the eVTOL’s maximum range. Additionally, the time losses associated with takeoff and landing are considered constant and represented by $TT_{\epsilon}$, which is independent of the strategy.

\subsection{Vertiport Waiting Time}

The waiting time at the departure vertiport $k_i$ is determined by the
discrete passenger queue length:
\begin{equation}
    TW^{\text{wait}}_{k_i}(t)
    =
    \frac{q^{\text{wait}}_{k_i}(t)}{\bar v^{\text{serv}}_{k_i}},
\end{equation}
where $q^{\text{wait}}_{k_i}(t) \in \mathbb{Z}_{\ge 0}$ denotes the number of waiting passengers at time step $t$. The queue length evolves according to:
\begin{equation}
    q^{\text{wait}}_{k_i}(t+1)
    =
    \max \Bigl\{
        q^{\text{wait}}_{k_i}(t)
        + N^{\text{arr}}_{k_i}(t)
        - N^{\text{dep}}_{k_i}(t)\, C_{k_i},
        \, 0
    \Bigr\},
\end{equation}
where $N^{\text{arr}}_{k_i}(t)$ is the number of newly arriving passengers,
$N^{\text{dep}}_{k_i}(t)$ is the number of eVTOL departures, and $C_{k_i}$ is
the passenger capacity of each eVTOL; $N^{\text{dep}}_{k_i}(t)=0$ indicates
that no eVTOL departs from vertiport $k_i$ at time step $t$. The $\max \{\cdot, 0\}$ operator ensures the non-negativity of the queue length. This formulation assumes a first-come-first-served (FCFS) discipline and deterministic service, which are commonly used in transportation queueing models to capture boarding delays at shared facilities.

For simplicity, the departure-based service can be approximated by an average
service rate
\begin{equation}
    \bar v^{\text{serv}}_{k_i}
    =
    \mathbb{E}\!\left[N^{\text{dep}}_{k_i}(t)\right] C_{k_i}.
\end{equation}

% Optimization Objective
\subsection{Optimization Objective}

The goal of the system is to assign an optimal pair of vertiports $(k_i, l_i) \in \mathcal{V} \times \mathcal{V}$ for each passenger $i \in \mathcal{N}$, such that the total travel time accumulated across all passengers is minimized. The optimization problem is formulated as:

\begin{equation}
    \min_{(k_i, l_i) \in \mathcal{V} \times \mathcal{V}} \sum_{i \in \mathcal{N}} TT^{\text{tot}}_{o_i, k_i, l_i, d_i}.
\end{equation}

This optimization seeks to balance passenger assignments across the network by jointly considering ground and air segments as well as vertiport queuing delays. This formulation allows for flexible and individualized vertiport routing decisions that collectively minimize the system-wide travel burden.

\begin{figure*}[!t]
	\centering
	\includegraphics[width=0.85\linewidth]{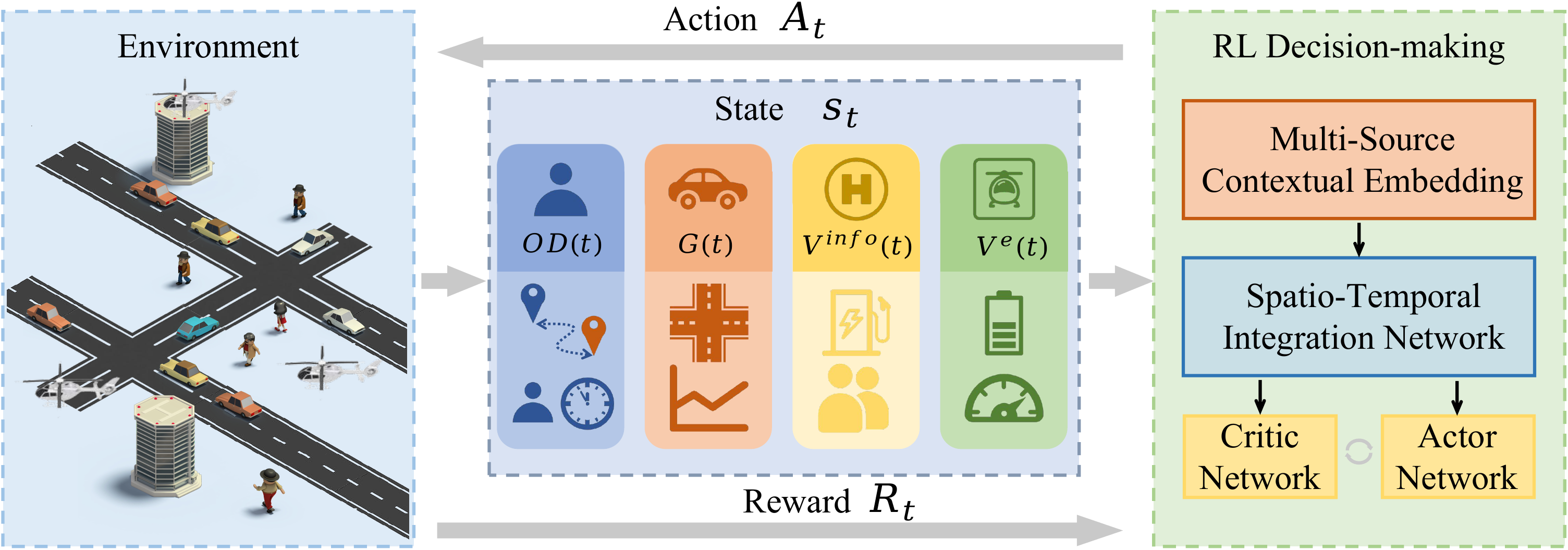}
	\caption{The proposed \textbf{U}nified \textbf{A}ir-\textbf{G}round \textbf{M}obility \textbf{C}oordination Optimization (UAGMC) framework.}
	\label{fig:model_design}
\end{figure*}

% Complexity Analysis
\subsection{Complexity Discussion}

The formulated vertiport assignment problem is combinatorially intractable, as each passenger $i \in \mathcal{N}$ can be assigned to any pair of distinct vertiports $(k_i, l_i) \in \mathcal{V} \times \mathcal{V}$, resulting in $|\mathcal{V}| (|\mathcal{V}| - 1)$ candidate combinations per passenger. When considering all passengers simultaneously, the solution space forms a high-dimensional assignment lattice, where the joint decision space grows rapidly with both the number of passengers and available vertiports. Consequently, the total number of feasible configurations across all passengers scales exponentially as $\mathcal{O} \left( |\mathcal{V}|^{|\mathcal{N}|} \cdot (|\mathcal{V}| - 1)^{|\mathcal{N}|} \right)$, making traditional combinatorial or enumeration-based optimization methods impractical for even moderately sized instances.

Beyond the combinatorial burden, the problem also exhibits strong temporal coupling due to evolving congestion levels and queuing dynamics at vertiports. Passenger decisions are inherently interdependent, as the assignments made by later passengers can increase waiting times and degrade travel efficiency for earlier passengers. For example, a passenger may choose a vertiport pair based on low initial congestion, only to experience delays as more passengers subsequently arrive at the same vertiport. This retroactive effect introduces non-stationary dynamics that complicate static optimization approaches. To overcome these challenges, we adopt a RL framework, which captures temporal interdependencies and system dynamics, providing a practical, data-driven solution to this complex assignment problem.

% Framework
\section{Proposed Method} \label{sec:framework}

This section presents the detailed design of the proposed framework. As illustrated in Fig.~\ref{fig:model_design}, the information required for decision-making is first collected from the environment. The \textbf{M}ulti-\textbf{S}ource \textbf{C}ontextual \textbf{E}mbedding (MSCE) processes the multimodal temporal scene information, encoding it into a series of latent feature vectors, which are then stored in the data buffer. Subsequently, the \textbf{S}patio-\textbf{T}emporal \textbf{I}ntegration \textbf{N}etwork (STIN) utilizes these latent vectors, considering temporal aspects, to extract relevant feature representations. Finally, a decision-making module based on Proximal Policy Optimization (PPO) generates decisions from the feature vectors produced by the STIN. In summary, the proposed RL framework supports effective decision-making under complex and heterogeneous state representations by incorporating passenger information, ground transportation conditions, and vertiport-related operational data.

\subsection{Markov Decision Process} 

The decision-making strategy for passenger travel can be regarded as a sequential decision-making problem, which can be formulated as an MDP. At each time step $t$, the system state $S(t)$ is observed via the V2X communication. Based on this observation, the agent makes availability and matching decisions according to a policy function $\pi(A(t) = a \mid S(t) = s)$. The system evaluates the policy through a scalar reward function defined as:
\begin{equation}
	R(t) = r(S(t), A(t), S(t+1)).
\end{equation}

The objective of the agent is to find an optimal policy $\pi^{*}$:
\begin{equation}
   \pi^{*}(s) 
     = 
     \arg\max_a \mathbb{E}\left[ \sum_{t=0}^{\infty} \gamma^t R(t) \mid s_0 = s, a_0 = a \right] ,
\end{equation}
where $ \gamma \in [0,1] $ is the discount factor that balances short-term and long-term rewards.

To elaborate on the components of this MDP, we define the state, action, and reward functions that are integral to the system's decision-making process.

% --> state <--
\subsubsection{State}
At time step $t$, the state input $S(t)$ is defined as follows:

\begin{equation}
  S(t) = \left[ OD(t); G(t); V^{\text{info}}(t); V^{e}(t) \right],  
\end{equation}

\begin{enumerate}[leftmargin=*]
    \item $OD(t)$: the request information that the control platform receives from passengers is represented as 
    $OD(t) = \{ (o^1_t, d^1_t), \ldots, (o^{|\mathcal{N}(t)|}_t, d^{|\mathcal{N}(t)|}_t) \} $, where $|\mathcal{N}(t)|$ denotes the number of passengers at time step $t$. Furthermore, $o^n_t$ and $d^n_t$ are the origin and the destination of the $n$-th passenger, respectively. 

    \item $G(t)$: $G(t) = \{ V_{g}(t); N^{v}(t) \}$ represents the ground traffic situation where $V_{g}(t) = \{ \bar{v}^{n}_{k_i} \}, n \in \mathcal{N}(t), k_i \in \mathcal{V} $ denotes the average speed of the $n$-th passenger traveling to the vertiport $k$. $N^{v}(t) = \{ n_1^v(t), \cdots, n_{|\mathcal{V}|}^v(t) \}$ denotes the number of passengers heading to vertiport $k$ at time $t$.

    \item $V^{\text{info}}(t)$: $V^{\text{info}}(t) = \{ Q^{\text{off}}(t), M^v, R^e \}$ represents the relevant information of vertiports at time~$t$, where $Q^{\text{off}}(t) = \{ q_1^{\text{off}}(t), \ldots, q_{|\mathcal{V}|}^{\text{off}}(t) \}$ indicates the number of passengers currently waiting at each vertiport; $M^v = \{ m_1^v, \ldots, m_{|\mathcal{V}|}^v \}$ represents the positions of the vertiports on the map; $R^e = \{ r_1^e, \ldots, r_{|\mathcal{V}|}^e \}$ denotes the charging rates of eVTOLs at each vertiport.

    \item $V^e(t)$: $V^e(t) = \{ (v_k^e(t), p_k^e(t), b_k^e(t)) \mid k \in \mathcal{V} \}$ represents the state of all eVTOLs at time $t$, including the flying speed, onboard passenger count, and remaining battery energy, where $v_k^e(t)$, $p_k^e(t)$, and $b_k^e(t)$ denote the average flying speed, onboard passenger count, and remaining battery energy of the $k$-th eVTOL at time $t$, respectively. The takeoff and landing time is omitted, as it is assumed to be relatively fixed and thus has a negligible impact on the decision-making process.   
\end{enumerate}

% --> action <--
\subsubsection{Action}
In our framework, the action $A(t)$ focuses on selecting departure vertiports for passengers, with landing vertiports assumed to be pre-determined by the system. This assumption leverages the long duration of aerial travel, which allows sufficient time to coordinate and assign landing vertiports, ensuring passengers can land at the vertiport closest to their destination. Therefore, the action space is defined as selecting the vertiport for the takeoff phase for each passenger in the air taxi.

The action $A(t)$ is defined as a vector of length $N(t)$. Each element of $A(t)$ represents the departure vertiport selected for an individual passenger:
\begin{equation}
    A(t) = [a_1, a_2, \dots, a_{N(t)}], \quad a_i \in \mathcal{V}, \, i \in \mathcal{N}(t),
\end{equation}
where $a_i$ is the index of the departure vertiport chosen for passenger $i$ at time $t$, and $\mathcal{V}$ denotes the set of feasible departure vertiports.

By restricting the decision-making to departure vertiports, the action design reduces computational complexity and avoids the need for dynamic optimization of landing assignments. This simplification ensures that computational resources are focused on optimizing vertiport utilization and minimizing waiting times at departure vertiports, which is critical for passenger satisfaction and system performance. Furthermore, assuming passengers land at the vertiport closest to their destination ensures that landing decisions have minimal impact on overall travel time.

% --> reward <--
\subsubsection{Reward}
The reward serves as the evaluation of actions taken by the agent in the environment. In supervised learning, it is typically used to provide explicit guidance on the actions the agent should take. However, in RL, the agent does not require direct guidance, which is one of the key advantages distinguishing RL from other learning paradigms. Nonetheless, designing an appropriate reward signal remains critical in RL, particularly in sparse reward settings, where the reward is zero most of the time and only provided for specific delayed events. In such cases, it becomes challenging for RL to associate a sequence of actions with delayed future rewards.

Our objective is to optimize the overall passenger experience in UAM-based air taxi services by minimizing the total travel time. Therefore, the average travel time across all passengers is used as the key metric for reward calculation, defined as:

\begin{equation} \label{ep:reward}
    R = \frac{\displaystyle\sum_{i \in \mathcal{N}} TT^{\text{tot}}_{o_i, k_i, l_i, d_i} }{ |\mathcal{N}|}.
\end{equation}

However, this reward signal is unavailable until passenger matching is performed and the total waiting cost for all passengers is determined. During this process, maintaining a buffer for incoming requests leads to zero reward. This sparse reward issue hinders RL from associating a sequence of actions with delayed rewards~\cite{hare2019dealing}.

To address this challenge, we propose a method to decompose the delayed reward into incremental rewards at each time step over the lifecycle of a passenger in the queue. Since the total matching waiting time can be expressed as the sum of waiting times at each time step, we define the incremental reward as the matching waiting reward:

\begin{equation} \label{ep:reward_incremental}
    R(t) 
    = 
    \frac{\displaystyle\sum_{i \in \mathcal{N}(t)} TT^{\text{tot}}_{o_i, k_i, l_i, d_i} }{|\mathcal{N}(t)|} 
    - 
    \frac{\displaystyle\sum_{i \in \mathcal{N}(t-1)} TT^{\text{tot}}_{o_i, k_i, l_i, d_i} }{|\mathcal{N}(t-1)|},  
\end{equation}
where $\mathcal{N}(t)$ denotes the set of all passengers who have arrived by time step $t$, including those who arrived in previous time steps. This approach effectively transforms the sparse reward problem into an incremental reward problem, thereby improving the learning efficiency of the RL agent.

\begin{figure}[htp]
	\centering
	\includegraphics[width=0.6\linewidth]{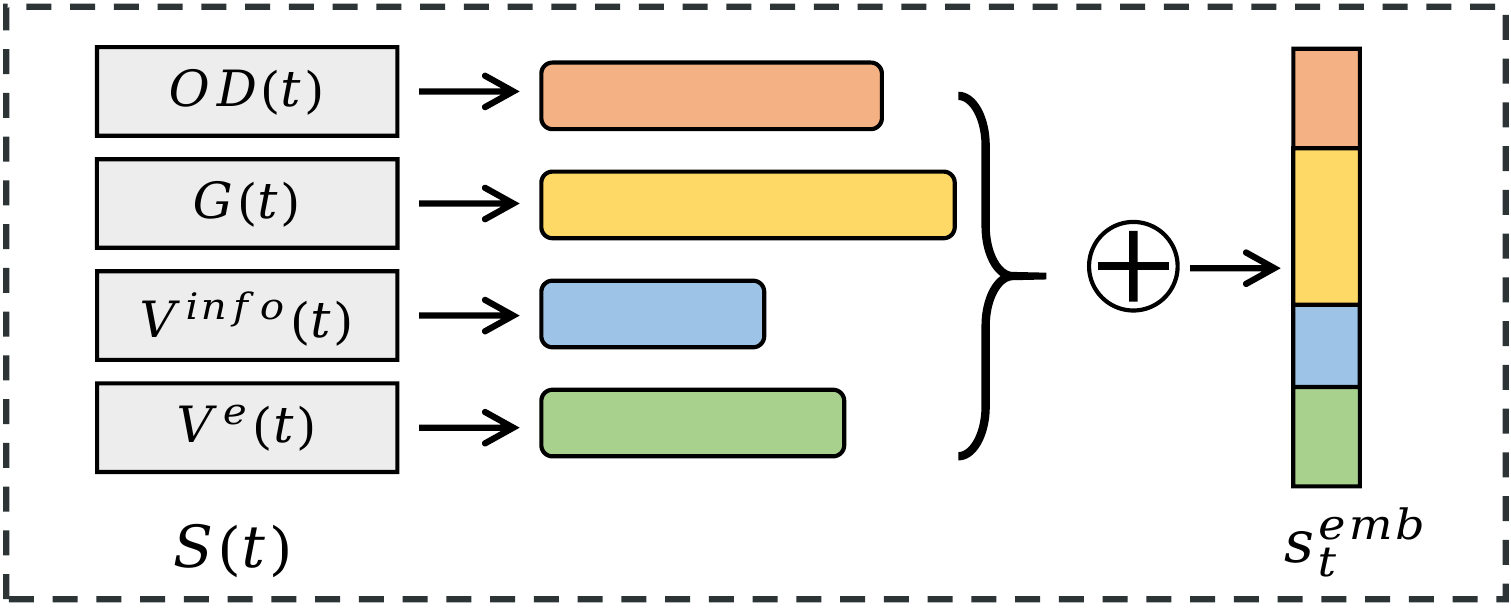}
	\caption{The \textbf{M}ulti-\textbf{S}ource \textbf{C}ontextual \textbf{E}mbedding architecture.} 
	\label{fig:msce}
\end{figure}

\subsection{Spatio-Contextual Feature Integration} % Air-Ground Mobility Representation
The integration of ground and aerial transportation systems forms a highly dynamic transportation network with an extremely large number of possible state-action pairs. Manually designed policies often fail to adapt to dynamic traffic changes and cannot comprehensively cover all possible states. Therefore, we employ \textbf{M}ulti-\textbf{S}ource \textbf{C}ontextual \textbf{E}mbedding (MSCE) and \textbf{S}patio-\textbf{T}emporal \textbf{I}ntegration \textbf{N}etwork (STIN) to process multi-source, high-dimensional, time-series raw observations and extract features, aiding the PPO model in more effectively handling the data.

The state features are multi-sourced, including dynamic, static, local, and global information. To better extract useful features from the high-dimensional heterogeneous state, we first encode the heterogeneous state information using MSCE, as illustrated in Fig.~\ref{fig:msce}.
\begin{equation}
    {\bm X}_t^{emb} = [s_{t-K+1}^{emb},s_{t-K+2}^{emb},\cdots,s_{t}^{emb}]
\label{eq:stacked_data}
\end{equation}
where
\begin{equation}
	s_j^{emb} = ReLU\left(emb({S}(j))\right)
	\label{eq:emb}
\end{equation}
for $j=t-K+1,t-K+2,\cdots,t$.

\begin{figure}[htp]
	\centering
	\includegraphics[width=0.5\linewidth]{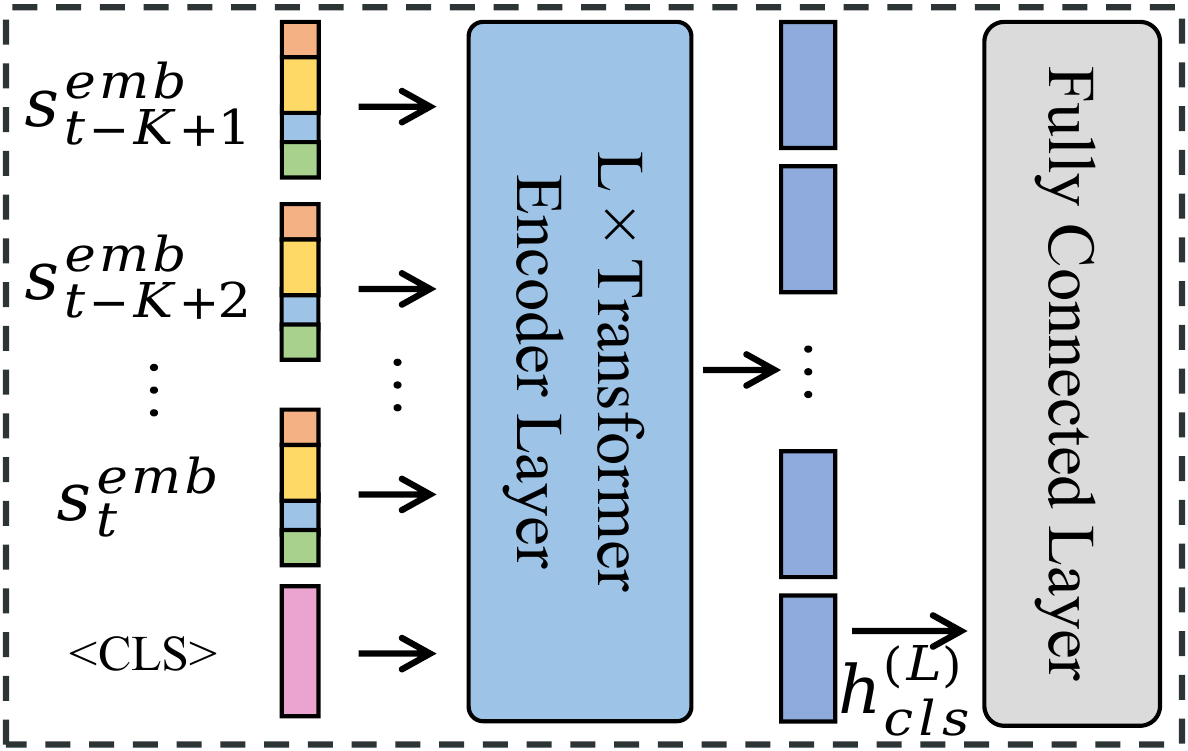}
	\caption{The \textbf{S}patio-\textbf{T}emporal \textbf{I}ntegration \textbf{N}etwork architecture, a multi-layer Transformer with a learnable CLS token and a fully connected layer to extract global temporal features.}
	\label{fig:stin}
\end{figure}

The objective is to optimize passenger travel strategies for the upcoming time interval by capturing the dynamic relationships between recent and current traffic conditions, thereby enabling informed decisions for future travel plans. To achieve this, we employ an attention-based Transformer as the STIN, which effectively processes heterogeneous time-series observational data. The architecture of this model is illustrated in Fig.~\ref{fig:stin}.

% 图片进行修改，其中 h^L_cls

The resulting sequential data ${\bm X}_t^{emb} \in \mathbb{R}^{K \times D}$ consist of the embedded state features $\{ s_j^{emb} \}$, as defined in Eq.~\eqref{eq:stacked_data}. A learnable classification token (CLS) $\mathbf{h}_{\mathrm{cls}} \in \mathbb{R}^{D}$ is prepended to the embedded sequence, forming the input to the Transformer encoder:
\begin{equation}
\tilde{\mathbf{X}}_t^{emb}
=
\left[
\mathbf{h}_{\mathrm{cls}},
\mathbf{s}_1^{emb},
\mathbf{s}_2^{emb},
\ldots,
\mathbf{s}_K^{emb}
\right]
\in \mathbb{R}^{(K+1)\times D}.
\end{equation}

After passing through multiple encoder layers, the output hidden representations are obtained as
\begin{equation}
\mathbf{H}^{(L)} = \left[ \mathbf{h}_{\mathrm{cls}}^{(L)}, \mathbf{h}_1^{(L)}, \ldots, \mathbf{h}_K^{(L)} \right],
\end{equation}
where $\mathbf{h}_{\mathrm{cls}}^{(L)}$ denotes the hidden state corresponding to the CLS token at the final layer.

The CLS representation is adopted as a global temporal feature summarizing the entire input sequence and is further projected by a fully connected layer to produce the final state embedding:
\begin{equation}
\mathbf{X}_t^{\mathrm{output}} = \mathbf{h}_{\mathrm{cls}}^{(L)} W_{\mathrm{fc}} + b_{\mathrm{fc}}.
\end{equation}

\subsection{Actor-Critic Algorithm}
In RL, the Actor-Critic algorithm is designed to separate the policy (actor) from the value function (critic). The actor selects actions based on the policy, while the critic evaluates these actions by estimating the value of the current state. This framework allows for simultaneous improvement of both action selection and value prediction. In this study, we adopt the PPO algorithm \cite{schulman2017proximal}, a variant of the Actor-Critic framework, to train the RL agent. PPO utilizes two neural networks: an actor-network $\pi_{\bm \Psi}$, responsible for determining the action distribution, and a critic network $V_{\bm \Phi}$, which estimates the expected return, both parameterized by ${\bm \Psi}$ and ${\bm \Phi}$. Input data is processed through a state encoding and STIN, generating a context vector ${\bm X}^{output}_{t}$, which informs both the actor's policy and the critic's value estimation. The PPO objective is optimized via a clipped surrogate objective $L^{CLIP}(\bm \Psi)$ to stabilize policy updates and improve reward acquisition, while a value function loss $L^{VF}(\bm \Phi)$ refines state value predictions. This approach addresses the latency challenges in traffic data acquisition and processing, enabling more efficient real-time decision-making. The overall loss function is given in the following form:
\begin{equation}\label{eq:overallloss}
    J({\bm \Psi},{\bm \Phi}) = \hat{\mathbb{E}}_t\left[L^{\text{CLIP}}({\bm \Psi}) - \lambda L^{\text{VF}}({\bm \Phi})\right],
\end{equation}
where $\lambda$ balances the two loss terms, and the PPO clip loss function $L^{\text{CLIP}}(\bm\Psi)$ can be represented as:
\begin{equation}
    L^{\text{CLIP}}(\bm\Psi) = \hat{\mathbb{E}}_t\left[\min(\rho_t(\bm\Psi){A}_t, \text{clip}\left(\rho_t(\bm\Psi), 1-\epsilon, 1+\epsilon){A}_t\right)\right],
\end{equation}
with 
\begin{equation}
    A_{t} = r_{t+1} + \gamma V({\bm X}^{output}_{t+1})-V({\bm X}^{output}_{t}).
\end{equation}

Finally, $\rho_{t}(\bm\Psi)$ denotes the ratio of the current strategy to the past strategy, as expressed in Eq.~\eqref{rt}.
\begin{equation} \label{rt}
    \rho_{t}(\bm\Psi)
    =
    \frac{\pi_{\bm\Psi}(a_{t} | {\bm X}^{output}_{t})}{ \pi_{{\bm\Psi}_{old}}(a_{t} | {\bm X}^{output}_{t})},
\end{equation}
where $\pi_{\bm\Psi}(a_{t} | {\bm X}^{output}_{t})$ and $\pi_{{\bm \Psi}_{old}}(a_{t} | {\bm X}^{output}_{t})$ denote the current and the past strategies. 

The value loss function $L^{VF}(\bm \Phi)$ in Eq.~\eqref{eq:overallloss} is typically calculated as the mean square error between the predicted state values $V_{\bm \Phi}({\bm X}^{output})$ and the actual discounted returns:
\begin{equation}
    L^{VF}({\bm \Phi}) = \frac{1}{N_{s}} \sum_{{\bm X}^{output}} (V_{\bm \Phi}({\bm X}^{output}) - \sum_{t=0}^{T} \gamma^t r_t)^2,
\end{equation}
where $N_s$ is the number of sampled states, $T$ is the time horizon, $\gamma$ is the discount factor, and $r_t$ is the reward received at timestep $t$. Minimizing this loss encourages the value network $V_{\bm \Phi}$ to accurately predict state values, which helps the agent estimate long-term rewards.

% %%%%%%%%%%%
% Experiment
% %%%%%%%%%%%
\section{Experiment} \label{sec:experiment}
This section presents the experimental design and key results. We begin by detailing the simulation environment. We then describe the comparative methods used for evaluation. Next, we present the metrics employed to assess system performance. Finally, we report the experimental results and conduct ablation studies to analyze the contribution of key components in our framework.

% Setup
\subsection{Simulation Setup}

% 补充 vertiport 的充电速度
To evaluate the proposed integrated air-ground scheduling method, we developed a simulation environment based on prior studies~\cite{bennaceur2022passenger, long2023demand, yedavalli2021assessing, fu2022scenario, wang2025transimhub}. In the simulation, three vertiports are deployed at designated coordinates, each representing distinct urban contexts. Following prior findings~\cite{yedavalli2022planning, straubinger2020overview, rajendran2021predicting}, passenger origin locations are dynamically generated to better capture the diversity of real-world travel demand. Among the vertiports, Vertiport~1 is located in a dense urban core, with surrounding passengers situated within a $0 \sim 6$ km radius, modeling typical inner-city access scenarios. In contrast, Vertiport~0 represents a low-density suburban or airport-access region, where passengers originate from within $6 \sim 12$ km. This urban-suburban configuration captures the spatial heterogeneity of UAM service areas and enables more realistic assessment of routing policies. The final site, Vertiport~2, is modeled as a pure landing destination without incurring any waiting time, serving as a simplified abstraction of urban sub-centers. Different vertiports are modeled with heterogeneous charging speeds and capacities to better reflect real-world variations in charging infrastructure and eVTOL handling capability. This design choice improves computational tractability while allowing the framework to focus on optimizing departure-side decisions.

To reflect realistic eVTOL operational practices, aircraft are dispatched only when all passenger seats are occupied, which maximizes vehicle utilization and mitigates excessive energy consumption caused by under-loaded flights. 
Such full-capacity dispatch assumptions are commonly adopted in UAM-related simulation studies to balance operational efficiency and system-level performance~\cite{yedavalli2021assessing, straubinger2020overview}. 
In addition, newly arrived eVTOLs at a vertiport are required to complete battery recharging before accepting subsequent passengers, capturing the practical charging and turnaround constraints observed in real-world eVTOL operations.

To simulate realistic vehicle dynamics, the mean cruising speed of eVTOLs is set to $120~\text{km/h}$, which lies within the typical operational range reported in recent survey and review studies on electric vertical take-off and landing aircraft~\cite{straubinger2020overview, xiang2024autonomous}. 
This value represents a conservative yet practical operating condition for short- to medium-range urban air mobility services.

For ground traffic, the maximum-flow speed is fixed at $60~\text{km/h}$~\cite{TrafficReport}, and speeds under varying densities are calculated using the Greenberg model, as described in Eq.~\eqref{eq:greenberg_model}. 
The RL agent is trained with a trace-decay factor $\gamma$ of $0.99$ and a clipping range of $0.2$ for policy updates. 
A summary of all simulation and training parameters is provided in Table~\ref{tab:simulation_variables}.

\begin{table}[!t]
	\centering
	\caption{Key simulation and RL training parameters.}
	\begin{tabular}{ccc}
		\toprule
		& \textbf{Variables} & \textbf{Value} \\ 
		\midrule
		\multirow{4}{*}{{\textbf{Environment}}} 
		& Map Size & $30 \, \text{km} \times 30 \, \text{km}$ \\  
        & Vertiport~0 Coordinates & $[0 \, \text{km}, 0 \, \text{km}]$ \\ 
        & Vertiport~1 Coordinates & $[10 \, \text{km}, 10 \, \text{km}]$ \\ 
        & Vertiport~2 Coordinates & $[27 \, \text{km}, 27 \, \text{km}]$ \\ 
        & Vertiport~0 Max eVTOL Capacity & $3$ \\ 
        & Vertiport~1 Max eVTOL Capacity & $1$ \\ 
        & Vertiport~0 Charging Speed & $800~\text{W}$ \\ 
        & Vertiport~1 Charging Speed & $300~\text{W}$ \\ 
		\midrule
		\multirow{5}{*}{{\textbf{Simulation}}}  
		& O-D Distance Range & $42 \, \text{km}$ to $48 \, \text{km}$ \\
		& CAV Max Speed $v_m$ & $60 \, \text{km/h}$ \\ 
		& eVTOL Cruise Speed $V^e$ & $120 \, \text{km/h}$ \\
        & eVTOL Seating Capacity $C^e$ & $3$ passengers \\
		& Passenger Demand $N(t)$ & $1$ \\ 
		& Time Horizon $T$ & $600$ time steps \\ 
		\midrule
		\multirow{4}{*}{{\textbf{RL Training}}} 
		& Trace-decay Parameter $\gamma$ & $0.99$ \\ 
		& Policy Update Parameter $\lambda$ & $1$ \\ 
		& Policy Clipping Range $\epsilon$ & $0.2$ \\ 
		& Length of State $K$ & $6$ \\ 
		\bottomrule
	\end{tabular}
	\label{tab:simulation_variables}
\end{table}

\begin{table}[!t]
	\centering
	\caption{Comparative results with six baseline methods. The $\pm$ symbol denotes the variance, representing the dispersion of individual passenger values from the mean.}
	\resizebox{\textwidth}{!}{\begin{tabular}{lcccccccc}
		\toprule
		\textbf{Model} & \textbf{ATT (min)} & \textbf{AET (min)} & \textbf{AWT (min)} & \textbf{AGT (min)} & \textbf{AAT (min)} & \textbf{RPV-0(\%)} & \textbf{RPV-1(\%)} \\
		\midrule
         Ground  & $59.9 \pm 3.4 $  & $59.9 \pm 3.4$ & -  & $59.9 \pm 3.4$ & - & - &- \\
		SPF & $133.65 \pm 2.37 $ & $27.12 \pm 0.03$ & $106.53 \pm 2.04$ & $5.93 \pm 0.01$ & $21.18 \pm 0.02$  & $55\%$ & $45\%$ \\ 
        Rule-Based & $62.30 \pm 6.72$  & $29.95 \pm 0.17$ & $32.34 \pm 5.23$ & $7.95 \pm 0.13$ & $22.01 \pm 0.07$  & $75\%$ & $25\%$ \\
        STTF & $52.39 \pm 1.36$ & $30.26 \pm 0.09$ & $22.13 \pm 1.41$ & $8.04 \pm 0.06$ & $22.22 \pm 0.03$ & $81\%$ & $19\%$ \\
        QTTI & $51.87 \pm 0.83$ & $30.26 \pm 0.09$ & $22.44 \pm 0.84$ & $7.23 \pm 0.04$ & $22.20 \pm 0.02$ & $80\%$ & $20\%$ \\
        Vanilla-PPO & $57.47 \pm 1.42$ & $30.30 \pm 0.05$ & $27.17 \pm 1.41$  & $8.12 \pm 0.05$   & $22.19 \pm 0.03$ & $80\%$ & $20\%$ \\
        UAGMC-L & $48.69 \pm 2.39$ & $30.72 \pm 0.08$ & $17.97 \pm 2.45$  & $8.51 \pm 0.05$   & $22.21 \pm 0.05$ & $80\%$ & $20\%$ \\
        \textbf{\textit{UAGMC-A}}  & $46.63 \pm 1.56$ & $31.15 \pm 0.04$ & $15.48 \pm 1.59$  & $8.83 \pm 0.03$   & $22.32 \pm 0.02$ & $83\%$ & $17\%$ \\
		\bottomrule
	\end{tabular}}
	\label{tab:evaluation_metrics}
\end{table}

\subsection{Comparison Methods}
To evaluate the effectiveness of our proposed UAGMC framework, we compare it against six representative baseline methods: (i) Ground: a fully ground-based strategy where all passengers travel exclusively via CAVs without using eVTOLs. (ii) Rule-Based: assigns passengers to vertiports based on predefined service rate ratios. (iii) SPF~\cite{archetti2022optimization}: selects routes that minimize physical travel distance. (iv) STTF~\cite{nie2009shortest}: optimizes routing based on estimated travel time under dynamic traffic conditions. (v) QTTI~\cite{zang2022travel}: integrates queuing delays and travel time into a unified cost function. (vi) Vanilla-PPO~\cite{schulman2017proximal}: a RL baseline with simple state concatenation, without explicit spatial-temporal modeling. (vii) UAGMC-L~\cite{pang2025v2x}: a variant of the UAGMC framework in which the STIN is implemented using an LSTM backbone. Our method, denoted as UAGMC-A, employs an attention mechanism in the STIN.

In addition to the aforementioned methods, we conduct ablation studies to investigate the individual contributions of the two proposed modules. These experiments enable us to isolate and analyze the effects of the MSCE and STIN modules on the learning performance of the deep RL framework.

% Metrics
\subsection{Evaluation Metrics}
We evaluate system performance using several metrics, each corresponding to different stages of the passenger journey. Given the stages, (1) ground travel to the departure vertiport, (2) waiting at the vertiport, (3) aerial travel, and (4) final ground travel to the destination, we select metrics that reflect both overall efficiency and component-wise performance.

To evaluate overall journey efficiency, we report the \textbf{Average Total Travel Time (ATT)}, which captures the complete travel duration across all four stages of a passenger's trip, including a fixed overhead for takeoff and landing operations. ATT can be decomposed into three key components: the \textbf{Average Ground Travel Time (AGT)}, which includes both the initial ground segment from origin to departure vertiport and the final segment from landing vertiport to the destination (Stages 1 and 4); the \textbf{Average Waiting Time (AWT)}, representing the time spent waiting at the departure vertiport prior to boarding (Stage 2); and the \textbf{Average Air Taxi Travel Time (AAT)}, measuring the in-flight duration between vertiports (Stage 3). In addition, to differentiate active travel from idle periods, we report the \textbf{Average Effective Travel Time (AET)}, which excludes AWT and reflects only the time spent in motion during Stages 1, 3, and 4.

In addition to time-based metrics, we analyze passenger distribution using \textbf{RPV-0} and \textbf{RPV-1}, which represent the proportion of passengers routed to Vertiport~0 and Vertiport~1, respectively. These ratios help evaluate spatial allocation and vertiport utilization under varying demand conditions.

\begin{figure*}[!t]
	\centering
	\subfigure[]{
		\includegraphics[width=0.45\linewidth]{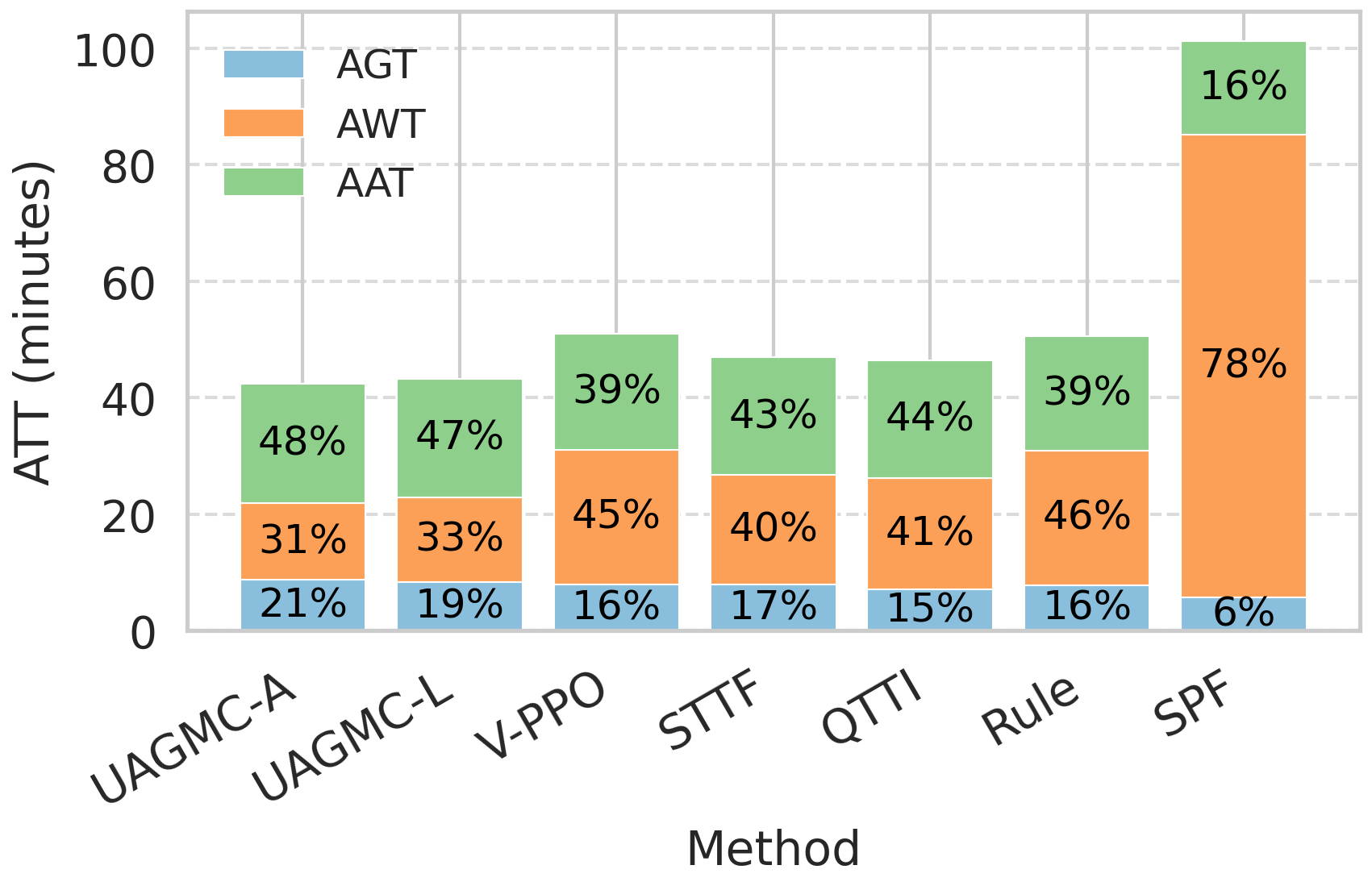}
		\label{fig:bar}
	} 
	\subfigure[]{
		\includegraphics[width=0.45\linewidth]{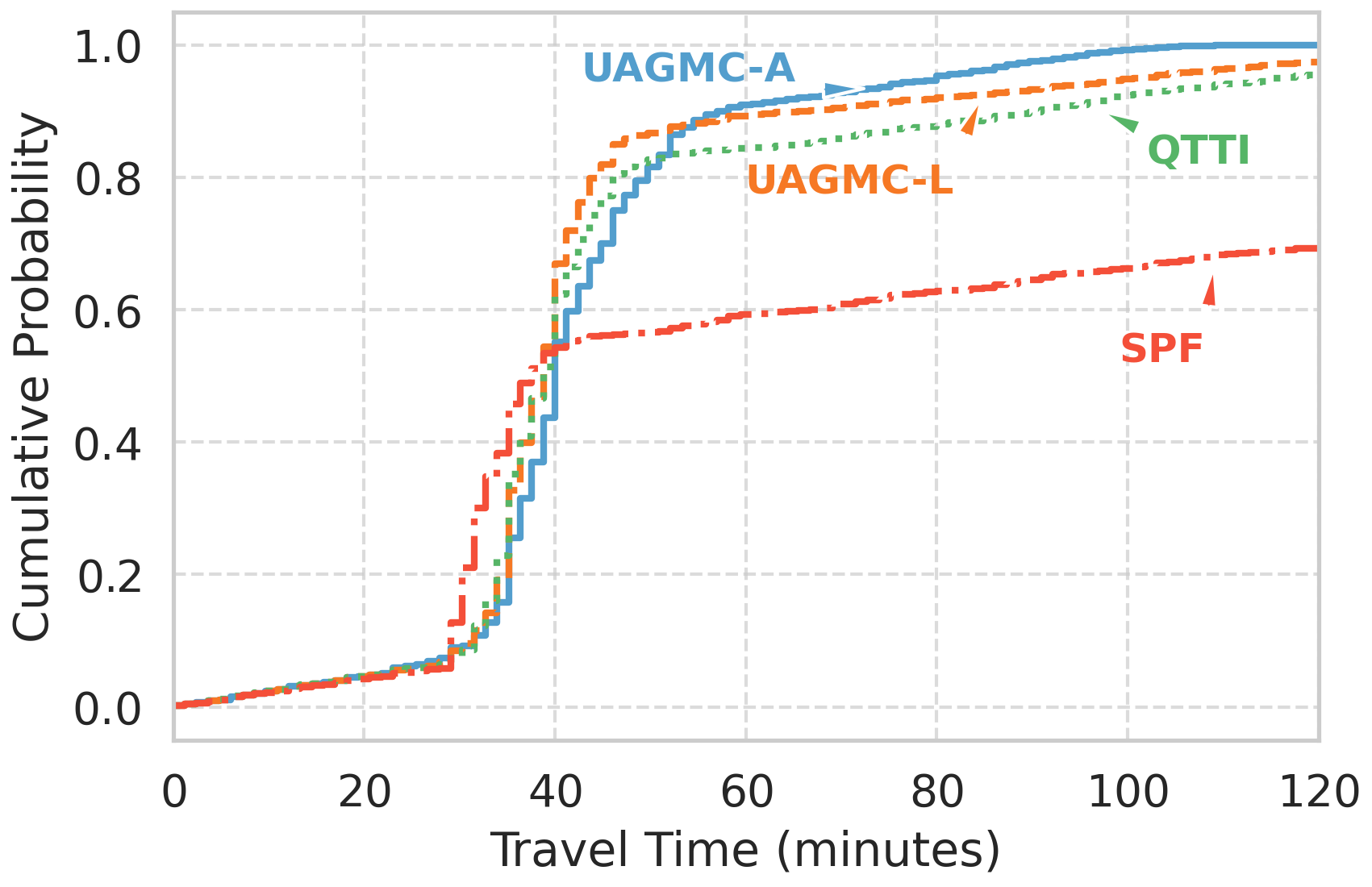}
		\label{fig:cdf}
	} 
	\caption{Comparative performance of baseline and proposed methods. (a) Decomposition of ATT into ground (AGT), waiting (AWT), and aerial segments (AAT). (b) Cumulative distribution of travel times highlighting system efficiency.}
	\label{fig:strategy_analysis_2}
\end{figure*}

\begin{figure*}[!t]
    \centering
    \subfigure[]{
        \includegraphics[width=0.45\linewidth]{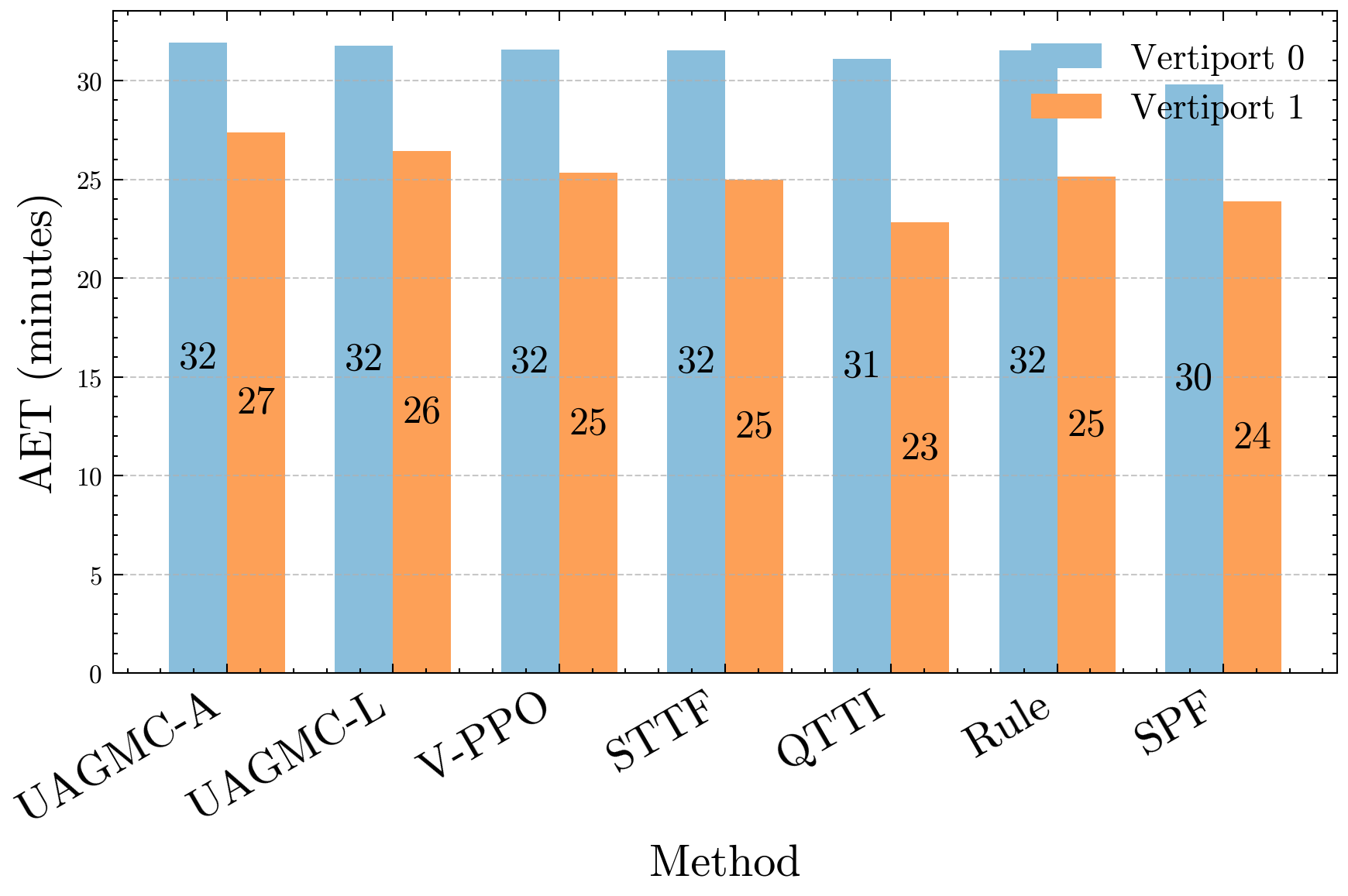}
        \label{fig:avg_travel_time}
    } 
    \subfigure[]{
        \includegraphics[width=0.45\linewidth]{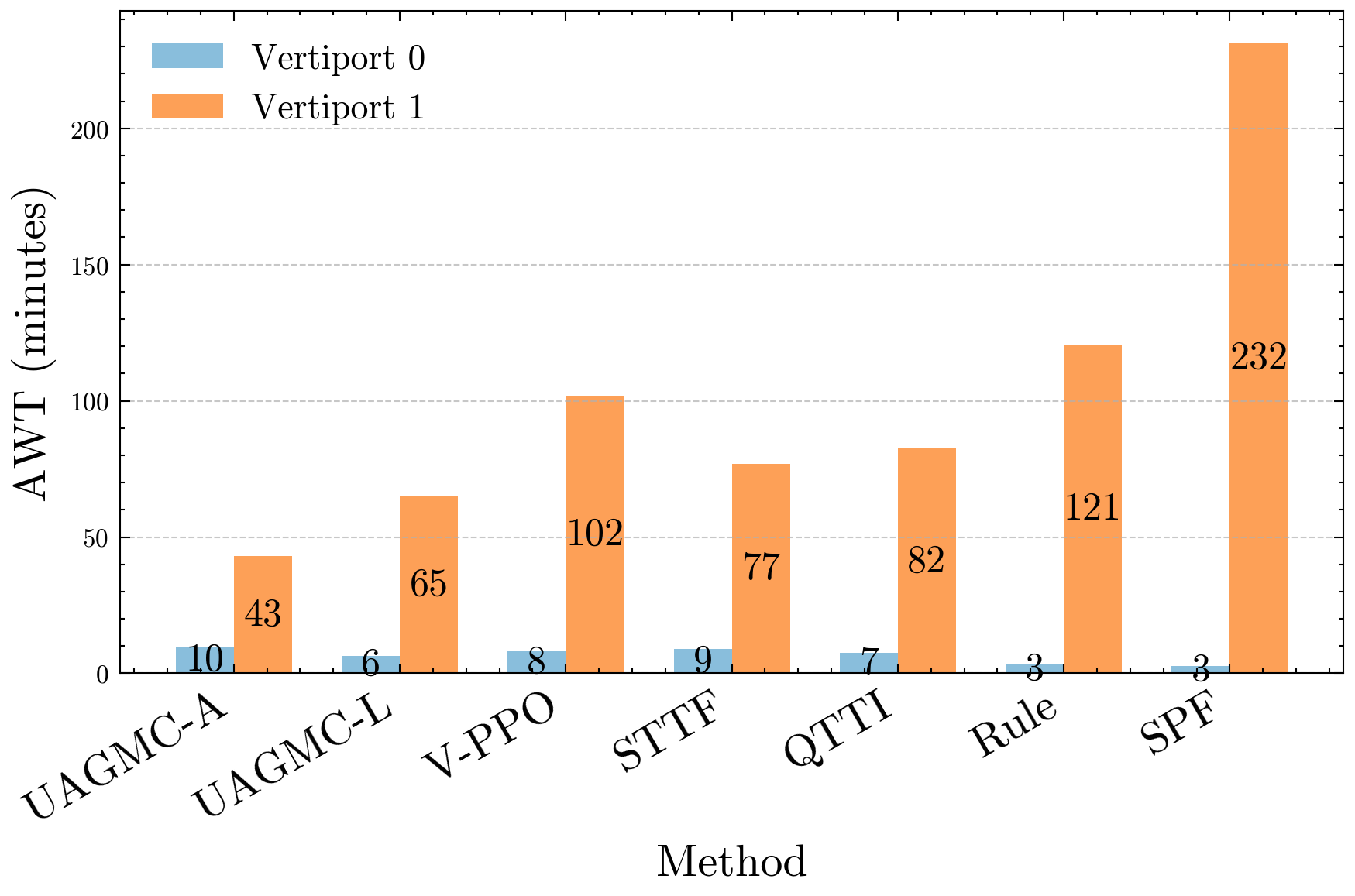}
        \label{fig:avg_wait_time}
    } 
    \caption{Travel time analysis for passengers assigned to Vertiport~0 and Vertiport~1 under different methods. (a) ATT (b) AWT}
    \label{fig:wait_travel_comparison}
\end{figure*}

\begin{figure*}[!t]
	\centering
	\subfigure[]{
		\includegraphics[width=0.45\linewidth]{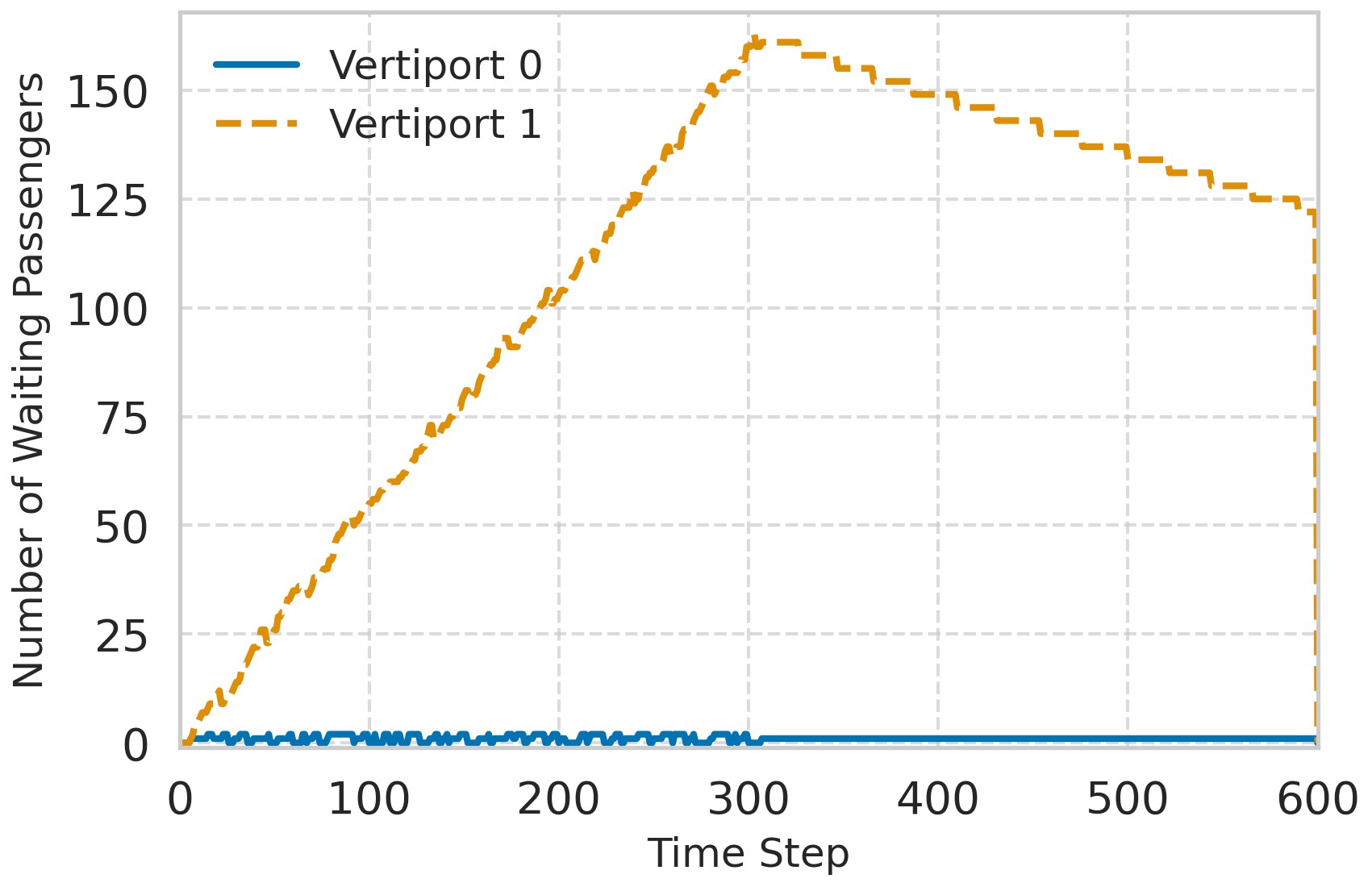}
		\label{fig:result_no_policy}
	} 
	\subfigure[]{
		\includegraphics[width=0.45\linewidth]{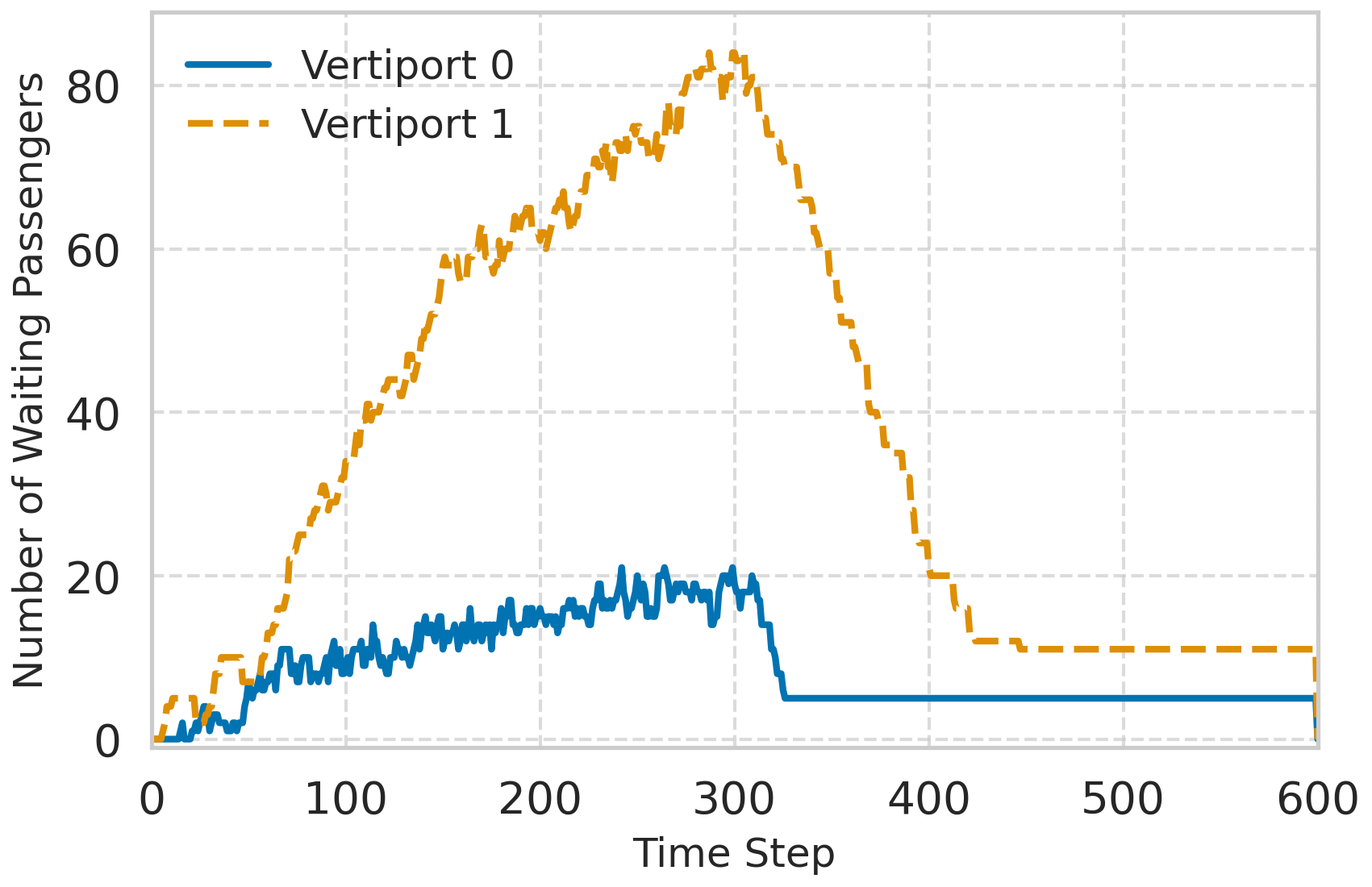}
		\label{fig:sttf_strategy}
	} 
	\subfigure[]{
		\includegraphics[width=0.45\linewidth]{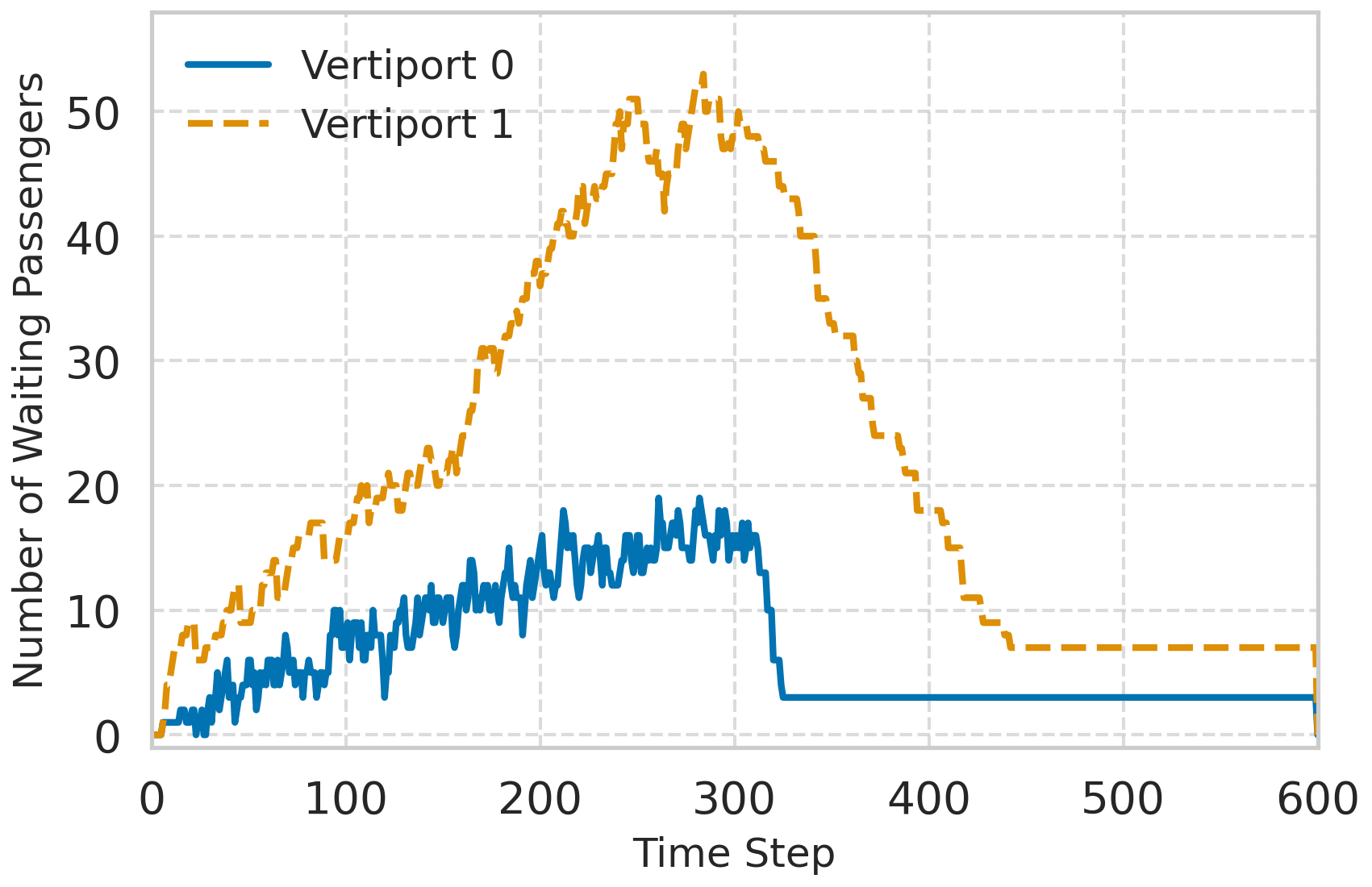}
		\label{fig:qtti_strategy}
	} 
	\subfigure[]{
	\includegraphics[width=0.45\linewidth]{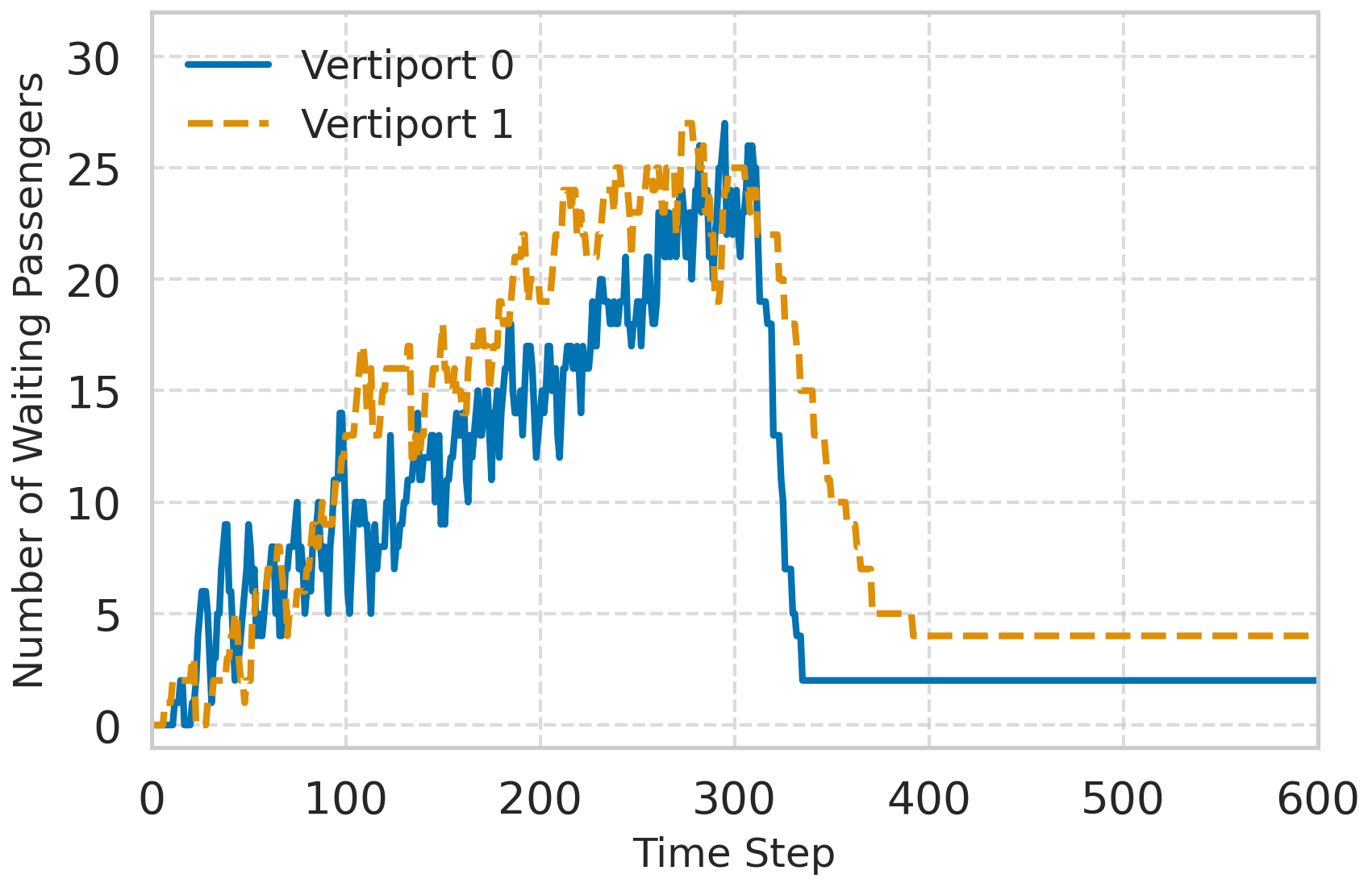}
	\label{fig:rl_strategy}
	}
	\caption{Passenger allocation strategies over time under different methods: (a) SPF, (b) STTF, (c) QTTI, and (d) UAGMC-A. The x-axis indicates simulation steps, and the y-axis shows the number of waiting passengers at each vertiport.}
	\label{fig:strategy_analysis}
\end{figure*}

\begin{table}[!t]
	\centering
	\caption{Model ablation experiments with MSCE and STIN. The $\pm$ symbol denotes the variance across different experimental runs.}
	\begin{tabular}{lccccc}
		\toprule
		& \textbf{MSCE} & \textbf{STIN} & \textbf{ATT (min)} & \textbf{AET (min)} & \textbf{AWT (min)} \\
        \toprule
		& \xmark & \xmark & $63.07 \pm 3.97$ & $30.25 \pm 0.39$ & $32.82 \pm 4.01$ \\
		& \checkmark & \xmark & $60.57 \pm 4.90$ &$30.49 \pm 0.09$ & $30.08 \pm 4.98$ \\
		& \xmark & \checkmark & $56.31 \pm 6.22$ & $30.74 \pm 0.14$ & $25.57 \pm 6.35$ \\
		\midrule
		& \checkmark & \checkmark &  $46.63 \pm 1.56$ & $31.15 \pm 0.04$ &$15.48 \pm 1.59$ \\
		\bottomrule
	\end{tabular}
	\label{tab:ablation}
\end{table}

\begin{table*}[!t]
\centering
\caption{Performance of \textit{UAGMC-A} under different passenger demand levels and eVTOL seating capacities. Results are averaged over 10 episodes. The $\pm$ symbol denotes the variance across passengers.}
%\label{tab:uagmc_a_param}
\begin{tabular}{ccccccc}
\toprule
\textbf{Capacity} &
\textbf{Passengers} &
\textbf{AWT(min)} &
\textbf{AET(min)} &
\textbf{ATT(min)} &
\textbf{RPV-0(\%)} &
\textbf{RPV-1(\%)} \\
\midrule
3 & 100 & $3.44 \pm 0.01$  & $29.55 \pm 0.05$ & $33.00 \pm 0.05$ & 73\% & 27\% \\
3 & 200 & $2.10 \pm 0.13$  & $30.22 \pm 0.08$ & $32.32 \pm 0.17$ & 84\% & 16\% \\
3 & 300 & $15.48 \pm 1.59$ & $31.15 \pm 0.04$ & $46.63 \pm 1.56$ & 83\% & 17\% \\
\midrule
4 & 100 & $5.31 \pm 0.12$  & $28.45 \pm 0.00$ & $33.76 \pm 0.12$ & 52\% & 48\% \\
4 & 200 & $3.50 \pm 0.24$  & $29.88 \pm 0.11$ & $33.38 \pm 0.28$ & 75\% & 25\% \\
4 & 300 & $2.00 \pm 0.21$  & $30.89 \pm 0.07$ & $32.88 \pm 0.19$ & 90\% & 10\% \\
\bottomrule
\end{tabular}
\label{tab:uagmc_a_param}
\end{table*}

% Results
\subsection{Experimental Results}

%% overall results
\subsubsection{Comparative Results with Baseline Methods}

Table~\ref{tab:evaluation_metrics} summarizes the overall performance of the six baseline methods and our proposed UAGMC framework, which includes two variants: one based on LSTM and the other based on an Attention mechanism. Both UAGMC variants outperform the baselines. Specifically, UAGMC-A achieves the lowest ATT of $46.63$ minutes, significantly improving over all baselines. Compared to the SPF method, which yields an ATT of $133.65$ minutes, UAGMC-A reduces ATT by approximately $65\%$. Even against the strongest rule baseline QTTI, UAGMC-A achieves a $10\%$ reduction in ATT, highlighting the benefit of dynamic and adaptive decision-making. The UAGMC-L variant~\cite{pang2025v2x} has been shown in previous work to achieve substantial improvements over the baselines. Building on this, our optimized UAGMC-A further enhances performance, achieving the lowest ATT of $46.63$ minutes, which corresponds to an additional reduction of approximately $4.3\%$ compared to UAGMC-L.

Fig.~\ref{fig:bar} decomposes the ATT into different stages, revealing that waiting time plays a critical role in overall performance. UAGMC-A achieves the lowest AWT of $15.5$ minutes, accounting for $31\%$ of the ATT, which identifies waiting delay as a major performance bottleneck. This improvement is attributed to UAGMC-A’s enhanced temporal and multi-dimensional feature extraction, enabling more accurate modeling of both in-transit and queued passenger dynamics. Consequently, UAGMC-A produces more informed vertiport assignments and better load balancing. In contrast, Vanilla-PPO relies on simple state concatenation without temporal modeling, leading to suboptimal allocations and longer waiting times.

Fig.~\ref{fig:cdf} presents the cumulative distribution function (CDF) of passenger travel times for all methods. The curve of UAGMC-A is relatively flat in the early stage but converges earliest overall, indicating stable and efficient performance across the entire system. In contrast, SPF exhibits a steep slope initially but flattens later, reflecting widely dispersed and generally longer travel times. These results highlight the robustness and effective scheduling capability of UAGMC-A in handling stochastic passenger flows.

Overall, UAGMC-A consistently outperforms rule-based and queuing-theoretic approaches by dynamically adjusting vertiport selection according to real-time state observations. Unlike STTF and QTTI, which rely on static cost heuristics or inaccurate queue predictions, UAGMC-A leverages learned representations to proactively mitigate congestion, thereby significantly reducing waiting times and improving end-to-end travel efficiency.

\subsubsection{Analysis of Vertiport Selection Strategies}

Differences in model performance largely stem from their passenger-to-vertiport assignment strategies. Fig.~\ref{fig:wait_travel_comparison} illustrates passenger distributions and the corresponding ATT and AWT values. UAGMC-A allows asymmetric assignment across vertiports to reflect differences in location and service capacity, avoiding artificially balanced but suboptimal routing decisions, unlike STTF which enforces near-equal ATT across vertiports.

UAGMC-A also achieves the lowest average waiting times at both vertiports, indicating that reductions in queuing delays are a primary driver of overall travel time improvement. In contrast, Vanilla-PPO can adjust allocations based on static capacities but lacks temporal modeling, resulting in slower responses to dynamic queue fluctuations and inferior performance.

Time-resolved analysis of passenger queues as shown in Fig.~\ref{fig:strategy_analysis}, further demonstrates the models' allocation behavior. SPF, STTF, and QTTI tend to favor the closer vertiport, reducing ground travel but causing frequent overloading and long queues. QTTI suffers from high variability due to inaccurate queue predictions. UAGMC-A, in contrast, maintains a more balanced distribution, dynamically redirecting passengers to avoid local congestion and stabilize overall system performance.

These results highlight a key trade-off in vertiport assignment: strategies that minimize access distance can inadvertently increase waiting times under capacity constraints, emphasizing the importance of jointly considering both spatial and temporal factors in multimodal UAM routing.

\subsection{Ablation Experiments}

To evaluate the contributions of the MSCE and STIN modules within the UAGMC framework, we conduct a series of ablation studies, as summarized in Table~\ref{tab:ablation}. When PPO is applied without either MSCE or STIN, the resulting performance is the worst among all variants, with an ATT of $63.1$ minutes and an AWT of $32.8$ minutes. This performance is even inferior to that of the single-frame Vanilla-PPO, indicating that simply extending the temporal span of observations, without a structured representation and effective temporal modeling, leads to high-dimensional and noisy inputs that hinder stable and efficient policy learning.

Incorporating MSCE, which embeds spatial and queue-related information into a structured representation, reduces ATT and AWT to $60.6$ and $30.1$ minutes, corresponding to improvements of $4.0\%$ and $8.2\%$ over the baseline. Using only STIN, which captures temporal dependencies, lowers ATT to $56.3$ minutes and AWT to $25.6$ minutes, yielding gains of $10.8\%$ and $22.0\%$. When both MSCE and STIN are combined, the model achieves the best performance, with ATT and AWT of $46.6$ and $15.5$ minutes, representing cumulative improvements of $26.1\%$ and $52.7\%$. These results demonstrate that integrating structured representation and temporal modeling is crucial for timely and effective decision-making under dynamic passenger flows.

\subsection{Parameter Sensitivity Analysis}
\label{sec:param_analysis}

Table~\ref{tab:uagmc_a_param} presents the performance of the UAGMC-A policy under different passenger demand levels and eVTOL seating capacities. As the total number of passengers increases, the system shows a clear transition from lightly loaded to congested operating regimes. For a fixed seating capacity of 3, the mean waiting time remains low at 100 and 200 passengers, but rises sharply at 300 passengers, indicating the onset of congestion. However, a sharp increase is observed at 300 passengers, indicating the onset of congestion. This effect is further reflected in the maximum waiting time, which grows rapidly under high demand, leading to a significant increase in the mean total travel time.

Increasing the eVTOL seating capacity to 4 effectively alleviates congestion effects. Both the mean and maximum waiting times are substantially reduced across all demand levels, and the mean total travel time remains stable even under heavy load. In contrast, the mean travel time exhibits limited variation across all configurations, suggesting that performance degradation is primarily driven by scheduling and queuing delays rather than changes in flight efficiency. Furthermore, the vertiport selection ratios adapt dynamically with system load, demonstrating the robustness of UAGMC-A in balancing spatial demand under varying operational conditions.

% %%%%%%%%%%%
% Conclusion
% %%%%%%%%%%%
\section{Conclusion} \label{sec:conclusion}

In this study, we addressed the problem of vertiport selection in on-demand air taxi services by formulating it as a mathematical optimization problem that integrates both ground and aerial transportation modes. To capture the complexity of multimodal travel under dynamic conditions, we reformulated the problem as a MDP and employed RL for solution generation. To handle heterogeneous information sources, such as ground traffic congestion and vertiport queuing status, we proposed the UAGMC framework, which incorporates a multi-source contextual embedding module and a spatio-temporal integration network. This design enables the model to process high-dimensional, multi-source data effectively, leading to more informed and adaptive routing decisions. Extensive experimental results demonstrate that UAGMC, equipped with an attention-based architecture, reduces the average total travel time by 34\%, significantly improving overall system efficiency compared to conventional methods. This work offers new insights into the integration of urban air mobility with ground transportation systems and provides a foundation for scalable, data-driven decision-making in future multimodal transport networks.

Although the proposed framework demonstrates strong performance, several critical aspects remain unaddressed, including the optimization and conflict management of vertiport landings, realistic airspace conditions, eVTOL route planning~\cite{xie2025real}, and the comprehensive scheduling of eVTOL operations across their entire missions. Future work will aim to tackle these challenges by incorporating multi-vehicle, end-to-end path planning, advanced conflict avoidance mechanisms, fine-grained operational management, and realistic communication constraints, thereby enhancing the model’s applicability and robustness in complex, real-world urban air mobility environments.

% %%%%%%%%%%
% Reference
% %%%%%%%%%%
\bibliographystyle{ieeetr}
\bibliography{refes}

\end{document}